\documentclass[twocolumn]{svjour3}

\usepackage{graphicx}%
\usepackage{multirow}%
\usepackage{amsmath,amssymb,amsfonts}%
\usepackage{mathrsfs}%
\usepackage[title]{appendix}%
\usepackage[dvipsnames]{xcolor}
\usepackage{textcomp}%
\usepackage{manyfoot}%
\usepackage{booktabs}%
\usepackage{algorithm}%
\usepackage{algorithmicx}%
\usepackage{algpseudocode}%
\usepackage{listings}%

\usepackage{subfigure}
\usepackage{pdflscape}
\usepackage{graphicx}
\usepackage{afterpage}

\usepackage{longtable}
\usepackage{multirow}
\usepackage{rotating}
\usepackage{tabularx}
\usepackage{threeparttable}    

\usepackage{hyperref}
\hypersetup{colorlinks=true,
linkcolor=blue,
citecolor=blue}

\usepackage[numbers]{natbib}

\DeclareMathOperator*{\argmin}{arg\,min}
\DeclareMathOperator*{\argmax}{arg\,max}

\begin{document}
\begin{sloppypar}

\title{A Survey on Global LiDAR Localization: Challenges, Advances and Open Problems}


\author{Huan Yin $^{1}$  \and
        Xuecheng Xu $^{2}$ \and
        Sha Lu $^{2}$ \and
        Xieyuanli Chen $^{3}$ \and \\
        Rong Xiong $^{2}$ \and
        Shaojie Shen $^{1}$ \and
        Cyrill Stachniss $^{4}$ \and
        Yue Wang $^{2}$
}

\authorrunning{Huan Yin \emph{et al.}} 

\institute{ Huan Yin (eehyin@ust.hk) \\
        Xuecheng Xu (xuechengxu@zju.edu.cn) \\
        Sha Lu (lusha@zju.edu.cn) \\
        Xieyuanli Chen (xieyuanli.chen@nudt.edu.cn) \\
        Rong Xiong (rxiong@zju.edu.cn) \\
        Shaojie Shen (eeshaojie@ust.hk) \\
        Cyrill Stachniss (cyrill.stachniss@igg.uni-bonn.de) \\
        Yue Wang (ywang24@zju.edu.cn) \\
       1 Hong Kong University of Science and Technology, Hong Kong SAR \\
       2 Zhejiang University, China \\
       3 National University of Defense Technology, China \\
       4 University of Bonn, Germany \\
 }


\maketitle

\begin{abstract}
Knowledge about the own pose is key for all mobile robot applications. Thus pose estimation is part of the core functionalities of mobile robots. Over the last two decades, LiDAR scanners have become the standard sensor for robot localization and mapping. This article aims to provide an overview of recent progress and advancements in LiDAR-based global localization. We begin by formulating the problem and exploring the application scope. We then present a review of the methodology, including recent advancements in several topics, such as maps, descriptor extraction, and cross-robot localization. The contents of the article are organized under three themes. The first theme concerns the combination of global place retrieval and local pose estimation. The second theme is upgrading single-shot measurements to sequential ones for sequential global localization. Finally, the third theme focuses on extending single-robot global localization to cross-robot localization in multi-robot systems. We conclude the survey with a discussion of open challenges and promising directions in global LiDAR localization. To our best knowledge, this is the first comprehensive survey on global LiDAR localization for mobile robots.


\end{abstract}


\tableofcontents

\section{Introduction}
\label{sec:intro}

Autonomous navigation is essential for a wide range of mobile robot applications, including self-driving vehicles on roads~\cite{liu2021role} and agricultural robots in farming~\cite{pretto2020building}. To achieve this, robot localization plays an indispensable role in virtually any navigation system. Today's tasks of mobile robots require these systems to operate in large-scale and constantly changing environments, posing potential challenges to robot localization and mapping.

The Global Navigation Satellite System (GNSS) is a widely used facility for robot navigation outdoors. GNSS facilitates robot localization primarily in two aspects. First, GNSS-fused methods can track the robot's \textit{local} motion continuously with limited error, such as GNSS-aided simultaneous localization and mapping (SLAM)~\cite{cao2022gvins}. The other underlying aspect is that GNSS can provide information on \textit{global} position. This information can help the robot initialize its position on Earth and recover its position if robot localization fails. In fact, both aspects are related to the two typical localization problems: \textit{pose tracking} and \textit{global localization}, which are introduced in the well-known Probabilistic Robotics~\cite{thrun2002probabilistic}. Unlike the pose tracking problem, global localization requires a robot to globally localize itself on a given map from scratch. Thus, the pose space is generally larger than that in the pose tracking problem, resulting in a challenging problem to solve.

GNSS heavily relies on the quality of data sent from satellites, making it impractical in GNSS-unfriendly areas, such as indoors, dense urban environments, or forests. In such environments, ultra-wideband (UWB) and other signal emitters~\cite{ito2014w} can be deployed for global localization. External markers and tags~\cite{olson2011apriltag} can also provide global position and orientation information for visual-aided localization. These methods rely on the distribution of external infrastructures, and modifying the environment is often not desirable. Hence, using onboard sensors without environment modification is a more general solution for mobile robots. Visual images are information-rich and easily obtained from cameras. Early approaches use cameras to achieve global visual localization~\cite{lowry2015visual}, which is a topic of significant relevance and has attracted lots of research interest~\cite{garg2021your}.

Light detection and ranging (LiDAR) sensors have seen significant development in the last 25 years. Early laser scanners only provided 2D laser points with low resolution and range~\cite{thrun2002probabilistic}. The development of sensor technology has propelled LiDAR sensing from 2D to 3D and from sparse to relatively dense point clouds. In the 2007 DARPA Urban Challenge, the Velodyne HDL-64E sensor was mounted on five of the six automated vehicles that completed the race~\cite{buehler2009darpa}. LiDAR sensors are now becoming standard equipment in the robotics community. LiDAR sensors directly provide distance measurements by emitting and receiving light. Compared to visual images from cameras, these long-range measurements are more robust to illumination and appearance changes, making global LiDAR localization more practical in large-scale and constantly changing environments. This motivates us to provide a comprehensive review of global localization using LiDAR sensors.

\subsection{Problem Formulation and Paper Organization}
\label{sec:problem}

Given a prior map $\mathbf{M}$ and input data $\mathbf{D}$, the estimation of robot states (poses) $\mathbf{X}$ can be formulated as follows using the Bayes rule,
{\small
\begin{equation} 
\centering
\hat{\mathbf{X}} = \argmax_{\mathbb{X}} p (\mathbf{X} \mid \mathbf{D}, \mathbf{M}) = \argmax_{\mathbb{X}} p(\mathbf{D} \mid \mathbf{X}, \mathbf{M}) p(\mathbf{X} \mid \mathbf{M})
\end{equation}}
in which $p(\mathbf{D} \mid \mathbf{X}, \mathbf{M})$ is the likelihood of given poses and map; $p(\mathbf{X} \mid \mathbf{M})$ is the prior information of $\mathbf{X}$. The map $\mathbf{M}$ is a critical factor for robot localization, and classical localization-oriented LiDAR maps are introduced in Section~\ref{sec:map} before surveying concrete localization methods. Specifically, the LiDAR maps are categorized into three types: keyframe-based submap in Section~\ref{sec:KM}, global feature map in Section~\ref{sec:GFM} and global metric map in Section~\ref{sec:GMM}.

For \textit{local pose tracking}, the prior distribution $p(\mathbf{X} \mid \mathbf{M})$ generally follows a specific unimodal distribution such as $p(\mathbf{X}) \sim \mathcal{N}(\cdot)$. However, for \textit{global localization}, the robot lacks knowledge of where it is, and the pose error can not be bounded. In the classical Probabilistic Robotics~\cite{thrun2002probabilistic}, the probability $p(\mathbf{X} \mid \mathbf{M})$ typically follows a uniform distribution without prior on the robot pose, i.e., $p(\mathbf{X} \mid \mathbf{M}) = \frac{1}{\left| \mathbb{X} \right|}$. The resulting estimation problem is given by:
\begin{equation}\label{eq:general}
\hat{\mathbf{X}} = \argmax_{\mathbb{X}} p(\mathbf{D} \mid \mathbf{X}, \mathbf{M})
\end{equation}
which is a general formulation of the global localization problem on a given map. The solution space is actually much larger than the local pose tracking problem, making it more challenging.

We start this problem with a single input and a single output. If $\mathbf{D}$ is a single LiDAR point cloud $\mathbf{z}_t$ at timestamp $t$, the problem is to estimate one global pose $\mathbf{x}_t$. This problem is referred to as a \emph{single-shot} global localization problem and is comprehensively reviewed in Section~\ref{sec:single} of this survey. The single-shot global localization problem can be formulated as a Maximum Likelihood Estimation (MLE) problem as follows:
\begin{equation}\label{eq:sing-shot}
\centering
\hat{\mathbf{x}}_t = \argmax_{\mathbb{X}} p(\mathbf{z}_t \mid \mathbf{x}_t, \mathbf{M})
\end{equation}

In~Section~\ref{sec:single}, we further classify single-shot methods based on the coupling degree of two different approaches: \textit{place recognition} and \textit{pose estimation}, which are two main categories of methods of this survey. Intuitively, place recognition achieves global localization in a retrieval manner while pose estimation provides a fine-grained metric pose. The coupling degree increases sequentially in Section~\ref{sec:PR}, Section~\ref{sec:PRPE}, Section~\ref{sec:PEPR}, and Section~\ref{sec:PE}. The processing of LiDAR measurement $\mathbf{z}_t$ and the form of the global map $\mathbf{M}$ also vary accordingly in these subsections. A detailed illustration is presented in Figure~\ref{fig:single-shot} and Section~\ref{sec:single}.

It is worth noting that the measurement $\mathbf{z}_t$ could be one collected LiDAR \textit{scan} at one timestamp~\cite{chen2021overlapnet} or one accumulated LiDAR \textit{submap} while the robot moves~\cite{dube2020segmap}. Both of them are described as LiDAR point clouds and can be considered as one measurement for a single-shot global localization system. We do not differentiate between these two measurements, although single-shot global localization with a sparse LiDAR scan is more challenging than with a dense LiDAR submap.

Typically, the size of a LiDAR map is much larger than that of a single LiDAR point cloud, i.e., $\left| \mathbf{M} \right| \textgreater \left| \mathbf{z}_t \right|$, making the single-shot problem hard to solve. To improve the performance of global localization, one direct approach is to use a continuous stream of scans or submaps as measurements, i.e., $\mathbf{D}=\mathbf{Z}_{t} \triangleq \left\{ \mathbf{z}_{k=1}, \cdots \mathbf{z}_t\right\}$. Then the original problem is converted to a \emph{sequential} global localization problem, which will be discussed in Section~\ref{sec:sequential} of this survey paper. The sequential global localization can be formulated as follows to estimate $\mathbf{X}_{t}$:
\begin{equation}\label{eq:sequential-batch}
\hat{\mathbf{X}}_t = \argmax_{\mathbb{X}} \prod^t_{k=1} p(\mathbf{z}_k \mid \mathbf{x}_k, \mathbf{M}) p(\mathbf{X}_{t})
\end{equation}
in which $p(\mathbf{X}_{t})$ contains the prior information, representing the connections of sequential $\mathbf{X}_{t}$. The estimation problem could be solved by fusing a sequence of single-shot global localization results in a \textit{batch processing} manner, similar to SeqSLAM~\cite{milford2012seqslam} for global visual localization. By solving this, global localization can provide a trajectory of robot poses relative to the map. Note that additional odometry information could help improve the sequential global localization by constraining the pose space~\cite{pepperell2014all}, and the input data is denoted as $\mathbf{D}=\left\{ \mathbf{Z}_{t}, \mathbf{U}_{t-1}\right\}$, where $\mathbf{U}$ denotes the odometry input of mobile robots.

However, in many practical applications, we may only be interested in the final global pose $\mathbf{x}_t$ with sequential input, for example, $\mathbf{x}_t$ as the initial guess for local pose tracking. On the other hand, the single-shot global localization result may not be so accurate and a back-end is desired to track multiple hypotheses. In this context, sequential global localization can be seen as a Markovian process to estimate $\mathbf{x}_t$, formulated as follows:
\begin{equation}\label{eq:sequential-markov}
\hat{\mathbf{x}}_t \propto \underbrace{p(\mathbf{z}_t \mid \mathbf{x}_t, \mathbf{M})}_{\text{Measurement}} \underbrace{p(\mathbf{x}_t \mid \mathbf{x}_{t-1}, \mathbf{u}_{t-1})}_{\text{Motion}}  \underbrace{p(\mathbf{X}_{t-1})}_{\text{Prior}}
\end{equation}
in which the measurement model and motion model are related to $\mathbf{z}_t$ and $\mathbf{u}_{t-1}$, and the prior $p(\mathbf{X}_{t-1})$ is determined by previous recursive inference. This formulation is also known as \textit{recursive filtering} for localization, and one representative work is Monte Carlo localization (MCL)~\cite{dellaert1999monte}. Both batch processing and recursive filtering here are the two main branches for robotic state estimation~\cite{barfoot2017state}.

\begin{figure*}[t]
  \centering
  \includegraphics[width=\linewidth,keepaspectratio]{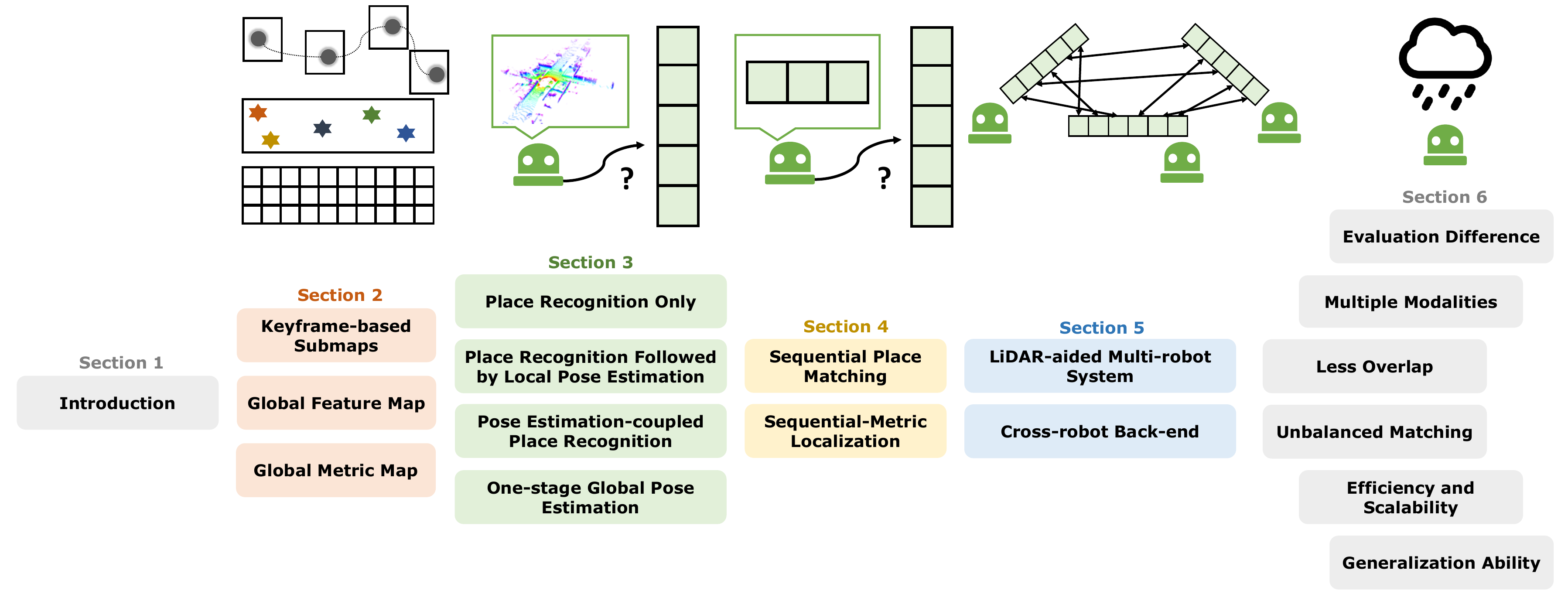}
  \caption{Fish-shaped paper structure. This survey starts the problem formulation and related introduction at the fish head. Then the fish body part contains the main subtopics of the global LiDAR localization problem: map framework, single-shot and sequential global localization, and cross-robot localization. Finally, an extended discussion on open problems is presented at the fish tail. We also present graphical illustrations above each section title.}
  \label{fig:framework}
\end{figure*}

As observed from the above equations, single-shot $p(\mathbf{z}_t \mid \mathbf{x}_t, \mathbf{M})$ still plays a key role in the sequential global localization problem. From another perspective, we can also categorize sequential global localization based on the use of place recognition and pose estimation, thus bridging the gap between Section~\ref{sec:single} and Section~\ref{sec:sequential}. We will introduce sequential place matching methods and sequential metric methods in Section~\ref{sec:seqMatch} and Section~\ref{sec:seqMetric}, respectively. The former mainly fuses sequential place recognition results and the latter focuses on estimating metric poses. Meanwhile, we also keep the discussion on the line of batch processing and recursive filtering in these two subsections.

Section~\ref{sec:single} and Section~\ref{sec:sequential} survey the mainstream methods for global LiDAR localization. In practical situations, global localization methods could not work very well in extreme conditions, such as localizing on an outdated map or localizing a robot on another robot's map. In Section~\ref{sec:cross}, we provide a review of several popular methods that could improve performance in such conditions, particularly focusing on multi-robot situations. Finally yet importantly, in Section~\ref{sec:discussion}, open problems of global LiDAR localization are discussed as a conclusion for future study.

In summary, our paper structure is similar to a fish, as illustrated in Figure~\ref{fig:framework}. Section~\ref{sec:intro} details the global localization problem and the scope of this survey. We then present three types of map frameworks in Section~\ref{sec:map}. Section~\ref{sec:single} and Section~\ref{sec:sequential} then provide an overview of existing methods based on the number of measurements: single-shot or sequential. The former focuses on matching a single LiDAR point cloud on a given map, while the latter takes sequential measurements to approximate the ground truth pose. Then in Section~\ref{sec:cross}, we extend the global localization problem to the cross-robot localization problem for multi-robot applications. Finally, Section~\ref{sec:discussion} provides discussions about open challenges and emerging issues of global LiDAR localization. A brief conclusion of this survey is presented in Section~\ref{sec:conclusion}.

\subsection{Typical Situations}
\label{sec:situations}

The concrete global localization method varies according to actual situations in robot mapping and localization. Three typical situations are illustrated as follows.

\subsubsection{Loop Closure Detection}
In a SLAM framework, loop closure detection (LCD) is a method used to determine whether a robot has returned to a previously visited location or place. However, simply recognizing revisited locations is insufficient for performing loop closure in SLAM. Typically, the relative transformation between the current and previous locations is also required, as is the case for graph-based consistent mapping methods~\cite{kummerle2011g,dellaert2012factor}. In this paper, we use the terms LCD and loop closing interchangeably, as both involve detecting revisited locations and estimating relative transformations. LCD is generally regarded as an \textit{intra-sequence} problem. The intra-sequence refers to scenarios where the measurements and map are derived from the same sequence, maintaining the continuity of the robot's journey. Conversely, the inter-sequence refers to the instance where the measurements and map are sourced from distinct data sequences, which could occur under varying time frames. These two terms were also introduced in the Wild-Places dataset~\cite{knights2022wild}.

\subsubsection{Re-localization} 
Re-localization serves to assist a robot in recovering when there is a failure in pose tracking or when the robot has been kidnapped. Additionally, it can be used to activate the robot at the beginning of navigation. The fundamental distinction between loop closure detection (LCD) and re-localization lies in the data sequence used: re-localization is classified as an \textit{inter-sequence} problem, wherein the measurements and map are obtained from different data sequences. It is noteworthy that re-localization can pose significant challenges in the case of long-term multi-session sequences, such as attempting to re-localize a LiDAR scan on an outdated point cloud map. Additionally, the pose space $\left| \mathbb{X} \right|$ of re-localization can be larger than that of LCD in some instances since there is no prior information available in re-localization, while LCD may employ odometry information as a crude initial estimate, thereby reducing the pose space to a smaller size.

\subsubsection{Cross-robot Localization}
Multiple online maps can be generated from multiple robots using incremental SLAM or other mapping techniques. These maps might have partial overlap but are under their own coordinate. Cross-robot localization, or multiple-robot mapping, aims to localize a robot globally on another robot's map. More concretely, $\mathbf{D}$ and $\mathbf{M}$ come from different robots and all robots' poses are required to be estimated. Theoretically, the cross-robot localization problem is identical to the single-robot re-localization~\cite{thrun2002probabilistic} but in a \textit{multi-robot} scenario. Relevant techniques can also be employed for offline map merging applications. For instance, cross-robot localization is performed on multiple sessions collected by a single robot for long-term use, while the challenge is that perspective changes may occur under long-term conditions.

\begin{figure}[t]
\centering
\includegraphics[width=\columnwidth]{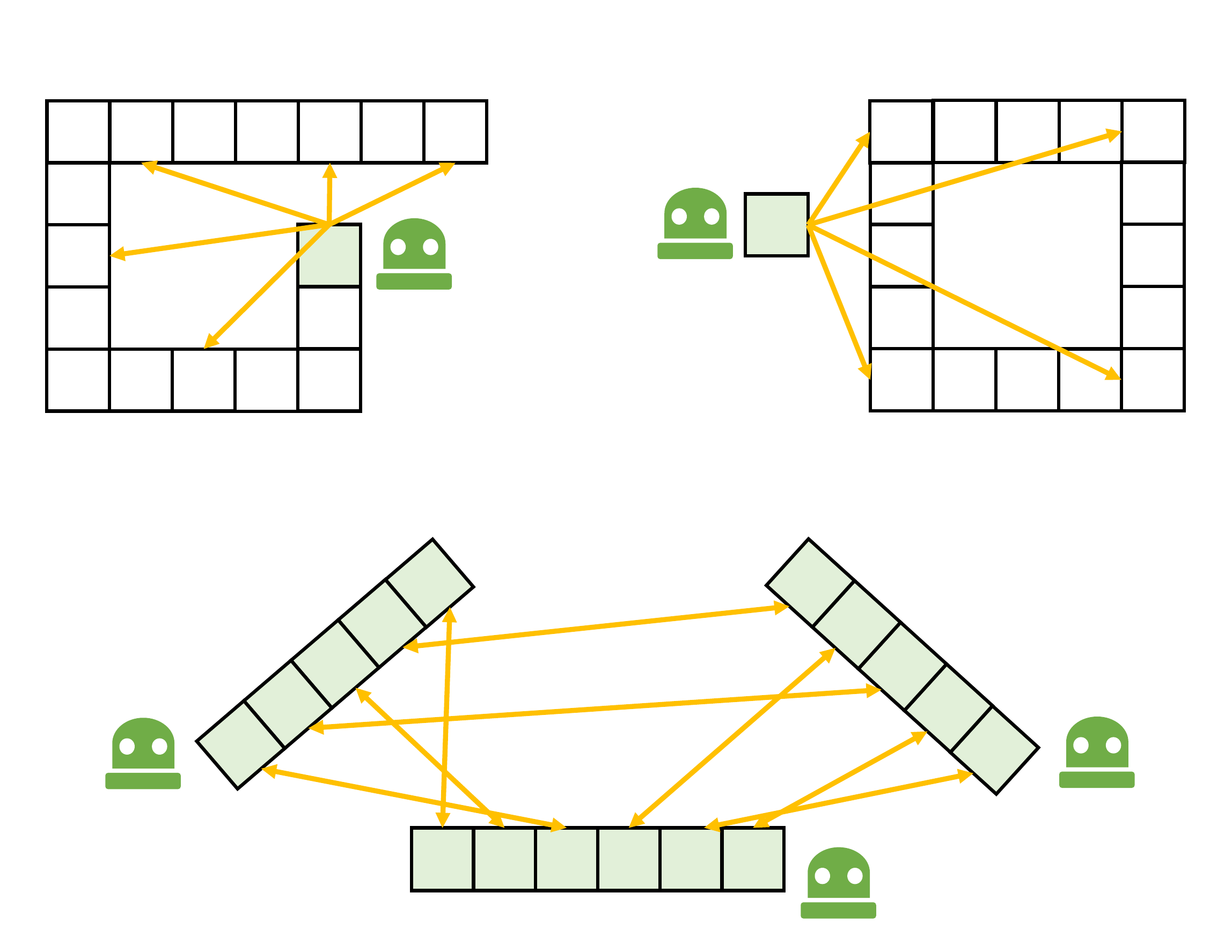}
\caption{Three typical situations. From top to down: single-robot intra-sequence LCD (loop closing); single-robot inter-sequence re-localization; cross-robot inter-sequence localization. Blue-filled boxes indicate measurements (LiDAR scan or submap). Orange lines are possible relative transformations for global localization problems.}
\label{fig:applications}
\end{figure}

Figure~\ref{fig:applications} illustrates three common scenarios in which a robot needs to estimate relative transformations between its current measurements and its own or another robot's map. This ability is commonly referred to as global localization, and it can be achieved with various sensors and fused sensor modalities. This survey specifically focuses on the global LiDAR localization problem and the techniques and open problems relevant to it. 

\subsection{Relationship to Previous Surveys}

Lowry \emph{et al.}~\cite{lowry2015visual} provide a thorough review on visual place recognition in 2015. They start by discussing the ``place'' definition and introduce related techniques for visual place recognition. A general place recognition survey~\cite{yin2022general} reviews the place recognition topic from multiple perspectives, including sensor modalities, challenges and datasets. However, place recognition determines whether a robot revisits a previous place by retrieval, which is not equal to the concept of global localization. Toft \textit{et al.}~\cite{toft2020long} review the long-term visual localization and make evaluations on state-of-the-art approaches, such as visual place recognition (image-retrieval)-based and structure-based camera pose estimation. Elhousni \textit{et al.}~\cite{elhousni2020survey} presents a LiDAR localization survey, focusing on LiDAR-aided pose tracking for autonomous vehicles. LiDAR place recognition and pose estimation are not reviewed explicitly in these survey papers~\cite{lowry2015visual,yin2022general,toft2020long,elhousni2020survey}. From the view of global LiDAR localization, we present a complete survey that covers relevant topics, like the ones~\cite{lowry2015visual,toft2020long} on vision. 

Cadena \emph{et al.}~\cite{cadena2016past} present a history of SLAM and promising research directions in 2016. SLAM has supported various robotic applications. A recent article by Ebadi \textit{et al.}~\cite{ebadi2022present} surveys recent progress on challenging underground SLAM. Specifically, SLAM aims to incrementally estimate pose and construct maps, while global localization estimates a global pose on a prior map. These two problems have a certain relevance. More concretely, LCD is a key characteristic of modern-day SLAM algorithms, as introduced in the Handbook of Robotics~\cite{stachniss2016simultaneous}. The absence of loop closing or place recognition will reduce SLAM to odometry~\cite{cadena2016past}. We believe this survey paper will help users make LiDAR SLAM systems more robust and accurate. 

\section{Maps for Global Localization}
\label{sec:map}

Before delving into the methodology section, it is essential to introduce maps $\mathbf{M}$ for robot localization. This section primarily focuses on maps that support global localization and classifies general-use maps into three primary clusters: keyframe-based submap, global feature map, and global metric map. We list three widely-used maps and discuss the map structure and representations inside.

\subsection{Keyframe-based Submap}
\label{sec:KM}

The keyframe-based submap is a highly popular map structure for robot localization, particularly in large-scale environments. It consists of a set of keyframes, each containing a robot pose and an aligned submap, as well as additional information in the form of topological or geometrical connections between keyframes~\cite{lowry2015visual}. Keyframe-based submaps are easy to maintain and well-suited for downstream navigation tasks~\cite{tang2019topological}. The keyframe-based map can be represented as:
\begin{equation} \label{eq:keyframe}
\mathbf{M}_{\text{sub}} = \left\{ \mathbf{m}_1, \cdots, \mathbf{m}_s \right\}
\end{equation}
where $s$ represents the number of submaps. In other words, $s$ corresponds to the size of $\mathbb{X}$ if we only retrieve places in the keyframe database.

A keyframe-based map effectively discretizes the entire pose space, reducing the complexity of the problem. This discrete map structure is particularly well-suited for place retrieval, as each keyframe can be considered a distinct "place" for the mobile robot. The submap contained within each keyframe can serve as a global descriptor for retrieval or can be augmented with additional metric grids or points for geometric registration. Notably, the distance between keyframe poses is a critical factor in practice. For example, if this distance is large or the keyframe resolution is low, fewer keyframes (i.e., a smaller $s$) may be needed for lightweight robot navigation, but at the cost of increased risk of localization failure. The content in each submap could be sparse features or dense metrics that will be introduced in the following sections. Additionally, it should be noted that keyframe-based maps may not be suitable for global localization in certain environments, such as indoor or forested areas where many local environments are similar. In such cases, a global map may be preferred.

\subsection{Global Feature Map}
\label{sec:GFM}

A global feature map keeps sparse local feature points to describe the environments. Early SLAM systems extract landmarks from laser data to support mapping and localization, like tree trunks in the Victoria Park dataset~\cite{guivant2001optimization}. These landmarks are essentially low-dimensional feature points. Nowadays, LiDAR feature points are generally with high-dimensional information~\cite{dube2017segmatch}. Hence, feature correspondence-based matching can be directly used for relative transformation estimation. More importantly, local features are sparse and easy to manage, making the navigation system more lightweight.

The main challenge of applying such maps is generating and maintaining stable feature points. For instance, a high-definition map (HD map) is a typical global feature map for self-driving vehicles. HD-map construction involves multiple onboard sensors and high-performance computation, and maintaining a global HD map is costly. As for the LiDAR-only global feature map, a powerful front-end feature extractor is necessary to ensure the map quality.

\subsection{Global Metric Map}
\label{sec:GMM}

A global metric map is a single map with dense metric representations describing a working environment. Generally, metric and explicit representations include 2D/3D points~\cite{lee2022learning}, grids~\cite{hess2016real},  voxels~\cite{wurm2010octomap}, and meshes~\cite{chen2021range}. The global metric map is easy to use and can provide high-precision geometric information.

But localization, whether pose tracking or global localization, is only one block in common autonomous navigation systems. In large-scale environments, the global metric map can be a burden for resource-constrained mobile robots. One might suggest that we could downsample or compress dense points while keeping the main geometric property~\cite{labussiere2020geometry,yin20203d}. But as pointed out by~\cite{chang2021map}, localization performances drop as the map size budget decreases using raw points. There are two solutions to tackle this problem: one is to use sparse local features rather than dense representations, i.e., the global feature map; another is to split the map space into submaps, i.e., keyframe-based submaps. The map framework and contents inside should be designed according to the application scenarios.

It is worth noting that implicit map representations are becoming highly popular, including non-learning~\cite{saarinen2013normal,wolcott2015fast} and learning-based ones~\cite{kuang2022ir,deng2023nerfloam}. One famous work is normal distribution transform (NDT), which uses probability density functions as representations. 
Current learning-based implicit representations~\cite{kuang2022ir,deng2023nerfloam} exploit techniques derived from neural radiance fields (NeRF)~\cite{mildenhall2021nerf}, which use fewer parameters compared with explicit ones~\cite{lee2022learning,hess2016real,wurm2010octomap,chen2021range} and have the potential to achieve higher accuracy owing to their continuous representations. Map representation is a basic but critical topic for SLAM and other navigation-related applications. We recommend reading a review by Rosen \textit{et al.}~\cite{rosen2021advances} for readers interested in this topic.

In summary, three types of maps are introduced in this section. These map structures and their representations inside are the foundations that can support global LiDAR localization in the following Section~\ref{sec:single} and Section~\ref{sec:sequential}. For instance, if place recognition technique is involved, like methods in Section~\ref{sec:PR}, \ref{sec:PRPE} and \ref{sec:PEPR}, space discretization is indispensable to obtain keyframe-based submaps for retrieval.

\begin{figure*}[t]
\centering
\subfigure[Section~\ref{sec:PR}: Place recognition only]{
	\includegraphics[width=0.45\linewidth]{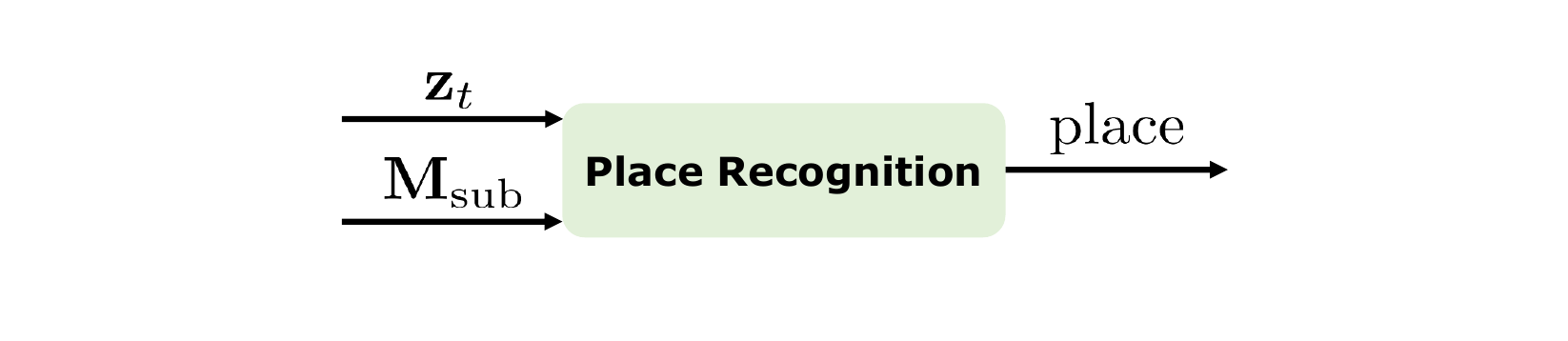}}
\subfigure[Section~\ref{sec:PRPE}: Place recognition followed by local pose estimation]{
	\includegraphics[width=0.45\linewidth]{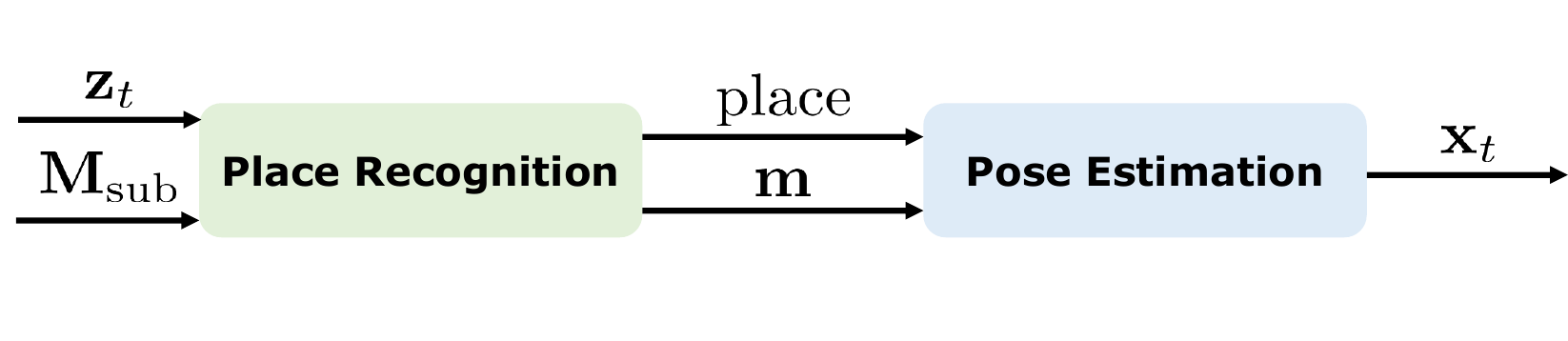}} \\
\subfigure[Section~\ref{sec:PEPR}: Pose estimation-coupled place recognition]{
	\includegraphics[width=0.45\linewidth]{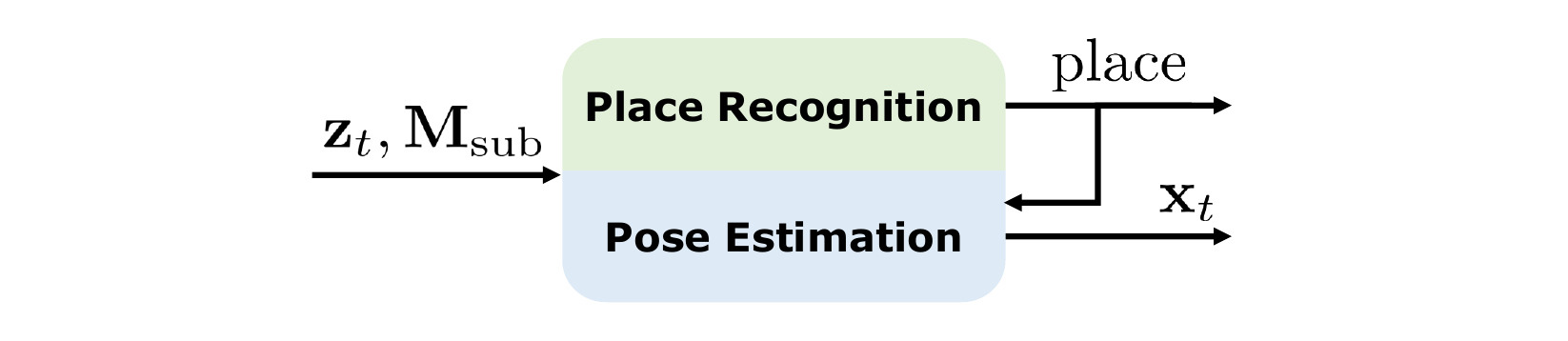}}
\subfigure[Section~\ref{sec:PE}: One-stage global pose estimation]{
	\includegraphics[width=0.45\linewidth]{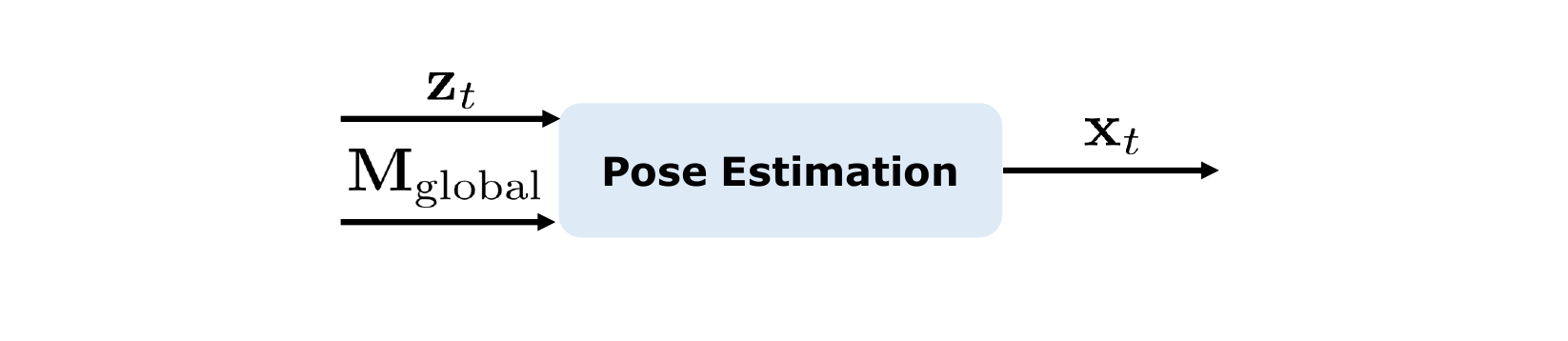}} 
\caption{Four types of single-shot global localization. The term $\mathbf{z}_t$ indicates the input LiDAR point cloud; $\mathbf{M}_{\text{sub}}$ and $\mathbf{M}_{\text{global}}$ represent keyframe-based submaps and a global feature map; \textbf{place} and $\mathbf{m}$ are the retrieved place and submap; $\mathbf{x}_{t}$ is the estimated pose. In Section~\ref{sec:PR}, place recognition-only approaches provide a retrieved place (keyframe) as the estimated pose. In Section \ref{sec:PRPE}, place recognition first provides a prior place then the pose is estimated via an individual pose estimation part. In Section \ref{sec:PEPR}, place recognition and pose estimation are coupled together and benefit from shared representations. Methods in Section \ref{sec:PE} achieve global pose estimation on a global map where place retrieval is not involved.}
\label{fig:single-shot}
\end{figure*}

\section{Single-shot Global Localization: \\Place Recognition and Pose Estimation}
\label{sec:single}


Single-shot global localization methods solve pose estimation using a \emph{single} LiDAR point cloud only. \textit{Place recognition} is the core backbone to achieve this. Generally, place recognition is a discriminative model based on keyframe-based submaps, in which every keyframe generally consists of a global descriptor and a robot pose. The basic idea of place recognition is to \textit{retrieve the highest-probability place based on global descriptors and measured similarities between} $\mathbf{z}_t$ and $\mathbf{M}_{\text{sub}}$. More specifically, these global descriptors should have a certain discriminativeness: be discriminative for different places but keep similar for places close to each other.

However, place recognition can only provide a coarse place as the estimated ``pose'', while local \textit{pose estimation} is still needed via precise feature matching or similar techniques. In this section, we categorize all single-shot approaches considering their degree of place recognition and relative pose estimation, as follows:

\begin{itemize}
    \item Section~\ref{sec:PR}: \emph{Place Recognition Only} approaches retrieve the most similar place using descriptors.
    \item Section~\ref{sec:PRPE}: \emph{Place Recognition Followed by Local Pose Estimation} first achieves place recognition and then estimates the robot pose via a customized pose estimator.
    \item Section~\ref{sec:PEPR}: \emph{Pose Estimation-coupled Place Recognition} tightly couple the two stages together.
    \item Section~\ref{sec:PE}: \emph{One-stage Global Pose Estimation} directly estimates the global pose on a global map using pose estimation. 
\end{itemize}

Figure~\ref{fig:single-shot} presents four types of combinations between the place recognition module and the pose estimation module. We also list several representative works of single-shot global LiDAR localization in Table~\ref{tab:single-shot}. From the perspective of maps, in Section~\ref{sec:PR}, \ref{sec:PRPE} and ~\ref{sec:PEPR}, methods generally rely on keyframe-based submaps. In Section~\ref{sec:PE}, global localization is generally based on a global feature map (or also with a metric map). Note that methods introduced in Section~\ref{sec:PRPE} focus on local pose estimation, and are applied when given place priors from methods in Section~\ref{sec:PR}.

The boundaries are not so clear for some global localization methods. For instance, there are no global descriptors in several place recognition approaches~\cite{bosse2009keypoint}, and local feature-based pose estimation plays an important role. We consider they lie in the boundary of Section~\ref{sec:PEPR} and Section~\ref{sec:PE}, and will list them in Section~\ref{sec:PEPR} for clearance.

{
\renewcommand{\arraystretch}{1.5} 
\begin{table*} \small
\centering
\caption{\textbf{Representative Studies in Section~\ref{sec:PR},\ref{sec:PRPE},~\ref{sec:PEPR} and~\ref{sec:PE}}}
\label{tab:single-shot}
\begin{tabularx}{\textwidth}{l l l l l X X  } 
\hline
Name & Place Prior & Pose & Pre-Processing  & Pipeline/Backbone/Highlight/Descriptor  \\
\hline
Fast Histogram \cite{rohling2015fast} & - & Place & Points  & Range Histogram-based     \\
PointNetVLAD \cite{uy2018pointnetvlad} & - & Place & Points & 3D PointNet + NetVLAD    \\
Minkloc3d \cite{komorowski2021minkloc3d} & - & Place & Points & 3D Feature Pyramid Network + Generalized-mean   \\
Kong \textit{et al.} \cite{kong2020semantic} & -& Place & Segments & Semantic Graph + Graph Similarity Network    \\
M2DP \cite{he2016m2dp} & - & Place & Projection & Multiple planes, Density Signature     \\
Yin \textit{et al.} \cite{yin2022fast} & - & Place & Projection & Spherical Projection + VLAD Layer    \\
\hline
FLIRT \cite{yang2013go} & \checkmark & 3-DoF & Points  & Curvature-based Detector + $\beta$-grid Descriptor  \\
TEASER \cite{TeaserTRO2021} & \checkmark & 6-DoF & Points & Invariant Measurements + Robust Estimation    \\
PHASER \cite{bernreiter2021phaser} & \checkmark & 6-DoF & Points & Spherical FT + Spatial FT + Correlations \\  
DPCN++ \cite{bernreiter2021phaser} & \checkmark & 6-DoF & Points & Differentiable Feature + FTs + Correlations \\
\hline
Scan Context \cite{kim2018scan} & - & Place + 3-DoF & Projection & 2D Ring Key, Scan Context     \\
DiSCO \cite{xu2021disco} & - & Place + 3-DoF & Projection & Multi-layer Scan Context + CNN + Fast FT       \\
OverlapNet \cite{chen2021overlapnet} & - & Place + 3-DoF & Projection & Multi-layer Range Image + CNN + Correlation       \\
Shan \textit{et al.} \cite{shan2021robust} & - & Place + 6-DoF & Projection & Range Image + DBoW + Matching     \\
LCDNet \cite{cattaneo2022lcdnet} & - & Place + 6-DoF & Points & PV-RCNN + BEV Feature Map     \\
STD \cite{yuan2022std} & - & Place + 6-DoF & Points & Stable Triangle Descriptor, Hash Key     \\
\hline
SegMatch \cite{dube2017segmatch} & - & 6-DoF & Segments & Features + Random Forest + RANSAC     \\
PointLoc \cite{wang2021pointloc} & - & 6-DoF & Points & PointNet-style + Self-Attention     \\
\hline
\end{tabularx}
\footnotesize \textbf{CNN: Convolutional Neural Network, FT: Fourier Transform}\\
\end{table*}
}

\subsection{Place Recognition Only}
\label{sec:PR}

\begin{figure}[t]
\centering
\includegraphics[width=\columnwidth]{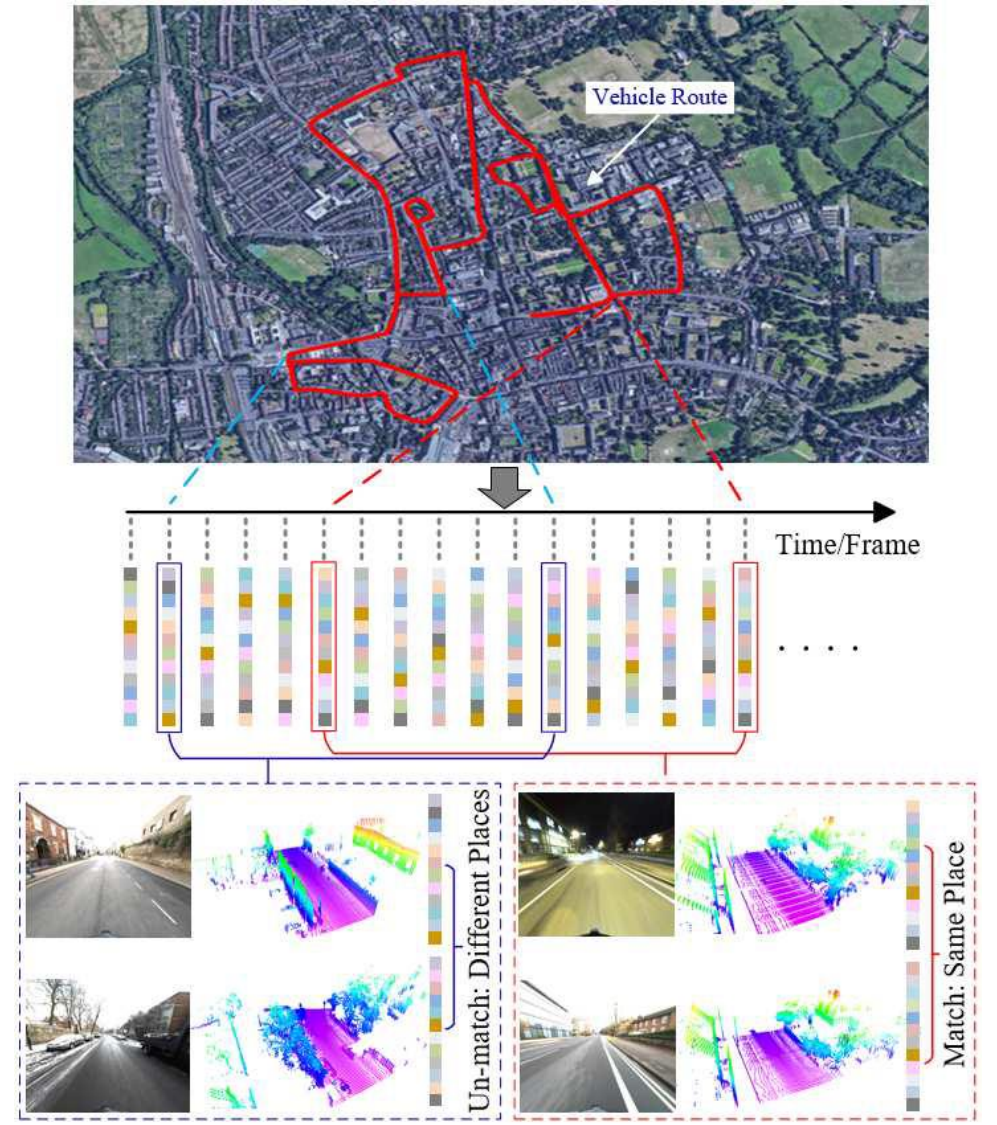}
\caption{LPD-Net is a place recognition-only approach for global LiDAR localization. Global descriptors are extracted as place descriptions for place retrieval. (Source: LPDNet~\cite{liu2019lpd}, used with permission.)}
\label{fig:lpdnet}
\end{figure}

Place recognition-only approaches solve the global localization problem by retrieving places in a pre-built keyframe-based map. Figure~\ref{fig:lpdnet} presents a place recognition-only approach for better understanding. The most challenging part of LiDAR place recognition is the global descriptor extraction. Compared to visual images, raw point clouds from LiDAR are textureless and in an irregular format, sometimes with an uneven density. From the perspective of data processing, global descriptor extraction is a kind of compression method for point clouds, while maintaining the distinctiveness of different places. We categorize place recognition based on how to handle LiDAR data pre-processing.

\subsubsection{Dense Points or Voxels-based} 
\label{sec:PR-dense}
Dense points and dense voxels-based works refer to those that generate global descriptors directly on dense representations. Early laser scanners can only provide 2D laser points for robotic localization. Granstr{\"o}m \textit{et al.}~\cite{granstrom2009learning} design a global descriptor that consists of 20 features in a 2D laser scan, such as a covered area and a number of clusters in range data. Then handcrafted descriptors and labels are fed into a weak classifier Adaboost~\cite{freund1997decision} for training. The learning-based approach is extended to 3D laser features in~\cite{granstrom2011learning}. Instead of extracting features, Fast Histogram~\cite{rohling2015fast} encodes the range distribution of 3D points into a one-dimensional histogram for place retrieval. Earth Mover’s distance is employed to measure the similarity of different histograms, which differs from Euclidean distance or Cosine distance in most place recognition methods. Inspired by~\cite{rohling2015fast}, Yin \textit{et al.}~\cite{yin2017efficient} build a 2D image-like representation based on divisions of altitude and range in a 3D LiDAR scan. Then the problem can be converted to an image classification problem that can be solved by training a 2D convolutional neural network with a basic contrastive loss~\cite{hadsell2006dimensionality}. Aside from using the range information of LiDAR scanners, DELIGHT~\cite{cop2018delight} utilizes the histograms of LiDAR intensity as the descriptor for place recognition followed by geometry verification.

All the methods above design handcrafted 2D or 1D histograms for LiDAR-based place recognition. This is because deep learning for 3D point clouds was not so mature then. In 2017, Qi \textit{et al.}~\cite{qi2017pointnet} propose the PointNet, which can learn local and global features for 3D deep learning tasks. Novel encoders also boost the performance of point cloud processing, like KPconv~\cite{thomas2019kpconv} for point convolution. PointNetVLAD~\cite{uy2018pointnetvlad} utilizes PointNet to extract features of 3D point clouds and aggregates them into a global descriptor via NetVLAD~\cite{arandjelovic2016netvlad}. But limited by PointNet, PointNetVLAD ignores the local geometry distribution in 3D point clouds. To address this problem, LPDNet~\cite{liu2019lpd} designs an adaptive local feature extraction module based on ten handcrafted local features, and a graph-based neighborhood aggregation module to generate a global descriptor. With the appearance of Transformer~\cite{vaswani2017attention} in diverse tasks to achieve long-range dependencies, the attention mechanism has been increasingly used to select significant local features for place recognition. PCAN~\cite{zhang2019pcan} takes local features into account and computes an attention map to determine each feature's significance. SOE-Net by Xia \emph{et al.}~\cite{xia2021soe} uses a point orientation encoding module to generate point-wise local features and feeds them into a self-attention network aggregating them to a global descriptor. Nevertheless, these methods cannot fully extract the point-wise local features around the neighbors. Hui \emph{et al.} propose a pyramid point cloud transformer network named PPT-Net~\cite{hui2021pyramid}. PPT-Net could learn the local features at different scales and aggregate them to a descriptive global representation by a pyramid VLAD. Recent work~\cite{linse} utilizes SE(3)-equivariant networks to learn global descriptors, making place recognition more robust to the rotation and translation changes. Despite the network structure design, a local consistency loss is proposed in~\cite{vidanapathirana2022logg3d} to guarantee the consistency of local features extracted from point clouds at the same place. To save memory and improve transmission efficiency, Wiesmann \textit{et al.}~\cite{wiesmann2022retriever} propose a compressed point cloud representation aggregated by an attention mechanism for place recognition. Authors also design a novel architecture for more efficient training and inference in~\cite{wiesmannkppr}.


Another popular pipeline is to voxelize the 3D point clouds first, and then extract global descriptors for place recognition. The voxeling process can make the raw 3D point clouds more regular. This makes 3D point clouds close to 3D image-like representations, i.e., each grid (2D) or cube (3D) can be regarded as one image patch. Magnusson \emph{et al.}~\cite{magnusson2009appearance,magnusson2009automatic} classify local cells into planes, lines and spheres, and then aggregate them all into a vector as a global descriptor for place recognition. The classification criteria are based on the local distributed probability density function, i.e., NDT. In the deep learning age, Zhou \emph{et al.}~\cite{zhou2021ndt} propose NDT-Transformer, which transforms the raw point cloud into NDT cells and uses the attention module to enhance the discrimination. VBRL proposed by Siva \emph{et al.}~\cite{siva2020voxel} introduces a voxel-based 3D representation that combines multi-modal features in a regularized optimization formulation. Oertel \emph{et al.} proposes AugNet~\cite{oertel2020augmenting}, an augmented image-based place recognition method that combines appearance and structure features. Komorowski \emph{et al.} introduce MinkLoc3D~\cite{komorowski2021minkloc3d}, which extracts local features on a sparse voxelized point cloud by feature pyramid network and aggregates them into a global descriptor by pooling operations. After that, they propose MinkLoc3Dv2~\cite{komorowski2022improving} as the enhancement of MinkLoc3D~\cite{komorowski2021minkloc3d}, which leverages deeper and wider network architecture with an improved training process.

\subsubsection{Sparse Segments-based} 
\label{sec:PR-segments}

Segmentation-based approaches refer to works that perform place recognition based on point segments, which leverage the advantages of both local and global representations. Seed~\cite{fan2020seed} segments the raw point cloud to segmented objects and encodes the topological information of these objects into the descriptor. SGPR~\cite{kong2020semantic} exploits both semantic and topological information of the raw point cloud and uses a graph neural network to generate the semantic graph representation. Locus~\cite{vidanapathirana2021locus} encodes the temporal and topological information to a global descriptor as a discriminative scene representation. Gong \textit{et al.}~\cite{gong2021two} utilize spatial relations of segments in both high-level descriptors search and low-level geometric search. Overall, segmentation-based approaches are close to what our human beings think about place recognition, i.e, using high-level representations rather than low-level geometry. On the other hand, these methods heavily rely on the segmentation quality and other additional semantic information. 3D point cloud segmentation approaches have typically been time-consuming and resource-intensive.

\subsubsection{Projection-based} 
\label{sec:PR-proj}

Projection-based methods, in contrast to the aforementioned two categories, do not generate the descriptor directly on 3D point clouds or segments; instead, these methods project a 3D point cloud to 2D planes first and then achieve global descriptor extraction. He at al.~\cite{he2016m2dp} propose M2DP that projects the raw point cloud into multiple 2D planes, constructing the signature with descriptors from different planes. LiDAR Iris~\cite{wang2020lidar} encodes the height information of a 3D point cloud into a binary LiDAR-Iris image and converts it into a Fourier domain to achieve rotation invariance. RINet proposed by Kong \emph{et al.}~\cite{li2022rinet} converts a point cloud to a scan context image encoded by semantic information first and designs a rotation-invariant network for learning a rotation-invariant representation. Yin \textit{et al.}~\cite{yin2022fast} design a multi-layer spherical projection via discrete 3D space. Then VLAD layer~\cite{arandjelovic2016netvlad} and spherical convolutions~\cite{cohen2018spherical} are integrated as SphereVLAD based on spherical projections. SphereVLAD could learn a viewpoint-invariant global descriptor for place recognition. 

\textbf{Summary.} Early approaches in Section \ref{sec:PR-dense} and \ref{sec:PR-proj} tried to design handcrafted global descriptors from a traditional data processing viewpoint. With the development of neural network techniques, data-driven descriptors are becoming more and more popular, resulting in high performance on place recognition (\textgreater95\% on Recall@1 in~\cite{xu2021transloc3d,komorowski2022improving}). Several approaches have achieved fully rotation-invariant descriptors for place retrieval, like handcrafted Fast Histogram in~\cite{rohling2015fast} and learning-based SphereVLAD in~\cite{yin2022fast}. We can conclude that global descriptor extraction of 3D LiDAR point clouds has reached a level of success. However, there still remain several challenges and issues, e.g., generalization ability, that will discuss in Section~\ref{sec:generalization}. 

All the methods in this subsection only provide retrieved places as output. The global localization performance is evaluated under machine learning metrics, like precision-recall curves and F1 score. We will discuss the evaluation metrics in Section~\ref{sec:evaluation}. In this context, the translation precision of pose (place) is decided by the resolution of keyframes (25m for evaluation on RobotCar Dataset~\cite{maddern20171}); the precision of rotation estimation is not considered or evaluated. In practice, this actually can not meet the demand of most high-precision global localization tasks, e.g., building a consistent global map with relative transformations, or waking the robot up with a precise location and orientation. 

From another point of view, global descriptors are highly compressed representations of raw LiDAR data, and there exists information loss in the compression process, especially for those end-to-end deep learning methods. This kind of representation is naturally suitable for nearest neighbor search in place retrieval but can not be used in geometric pose estimation. In the following section, we present a review of the local transformation estimation that metric representation involves.

\subsection{Place Recognition Followed by Local Pose Estimation}
\label{sec:PRPE}

This section reviews local pose estimation methods for high-precision transformation estimation. Note that this local pose estimation is independent of place recognition in this subsection. These two components are seen as separated and the global localization is achieved in a coarse-to-fine manner: first achieve place retrieval on keyframe-based submaps, then apply local pose estimation via matching input LiDAR to map data attached on the retrieved keyframe. Hence, for this group of approaches, the keyframe includes not only global descriptors for nearest neighbor search (place retrieval), but also metric representations for local pose estimation. Conventionally, the local pose estimation is achieved by precise point cloud registration.

Point cloud registration, or named scan matching, is a popular topic in robotics and computer vision. It aims at estimating the optimal transformation by minimizing the error function as follows:
\begin{equation} \label{eq:pointreg}
\mathbf{T} = \argmin_{\mathbf{T}\in\text{SE(3)}} \left( e\left(\mathcal{M}, \mathbf{T}\mathcal{P}\right) \right),
\end{equation}
in which $\mathbf{T}$ is the relative transformation (pose) to be estimated; $\mathcal{P}$ and $\mathcal{M}$ are the source points (input LiDAR measurement $\mathbf{z}_t$) and target points (prior map in the retrieved keyframe $\mathbf{m}$) respectively; $e\left( \cdot\right)$ is an error function to minimize. Specifically, point cloud registration approaches can be categorized into two types based on whether they use explicit correspondences in space for pose estimation, i.e., correspondence-based and correspondence-free approaches. This subsection mainly focuses on the \textit{global} point cloud registration, i.e., aligning two LiDAR point clouds without initial guess.

\subsubsection{Correspondence-based}
\label{sec:PRPE-Correspondence}

If correspondences (data associations) between query measurements and map are known, the registration problem can be solved in a closed form~\cite{horn1987closed,arun1987least}. Unfortunately, the initial correspondences are unknown in practice. The most well-known algorithm to scan registration is Iterative Closest Point (ICP)~\cite{besl1992method}, which considers a basic point-to-point correspondence search and finds the optimal solution at each iteration. The ICP family follows an expectation-maximization framework that alternates between finding correspondence and optimizing pose. Despite its widespread use in point cloud registration, the quality of the registration result is limited by the presence of noise and outliers. An effective real-time registration system based on ICP is KISS-ICP~\cite{vizzo2022kiss}. To improve the original ICP algorithm, many variants have been designed. Probabilistic methods Generalized-ICP~\cite{segal2009generalized} and NDT~\cite{biber2003normal} define Gaussian models for points or voxels and perform registration in a distribution-to-distribution manner, therefore reducing the influence of noise. We recommend interested readers consider a registration review for mobile robotics~\cite{pomerleau2015review}.

However, ICP and its variants might fall into local minima, making it inapplicable for global registration. Go-ICP by Yang \textit{et al.}~\cite{yang2013go} provides a global solution to the registration problem defined by ICP in 3D using branch-and-bound (BnB) theory. Go-ICP, however, is time-consuming on resource-constrained platforms, especially when the pose space is large for BnB search. If the transformation is in a limited space, BnB-based scan matching is more efficient to use, like LCD in Cartographer~\cite{hess2016real} and vehicular pose tracking on a Gaussian mixture maps~\cite{wolcott2015fast}.


For ICP and its variants, the local minima are caused by the assumption of nearest-neighbor correspondence in Euclidean space. Local feature-based approaches have emerged to extract robust features for correspondence search in a feature space. With the correspondence determined, the transformation can be calculated in the closed form, or with an additional outlier filter. But compared to 2D image descriptors like SIFT~\cite{lowe1999object} or ORB~\cite{rublee2011orb}, the study on LiDAR feature extraction and description is less extensive. The nature of range data is different from image data. Extracting and describing repeatable features in LiDAR scans is still an open problem. The less accurate correspondences provided by feature matching will cause a much higher outlier rate than their 2D counterparts. To address these issues, there are mainly two lines of research in recent years: one is to study the effective LiDAR features; the other is to configure a robust estimator that can handle high outlier rates. We will address these two lines as follows.

The feature extraction of 2D laser scans follows the pipeline in computer vision: first detect interest points (keypoints), and then compute a distinctive signature for each of them (local descriptors)~\cite{nielsen2022survey}. Tipaldi and Arras~\cite{tipaldi2010flirt} propose a fast laser interest region transform
(FLIRT) fo feature extraction, which adopts the theory in SIFT~\cite{lowe1999object}. FALKO~\cite{kallasi2016fast} is also an effective keypoint detection that specialized in 2D range data. BID by Usman \textit{et al.}~\cite{usman2019extensive} uses B-spline to fit the data along keypoints that are detected by FALKO~\cite{kallasi2016fast}. Then the spline is formulated into the descriptors for feature matching. As for 3D range data, early approaches to extract 3D features are mainly handcrafted~\cite{guo2016comprehensive}, such as FPFH~\cite{fpfh2009}, NARF~\cite{steder2010narf} and SHOT~\cite{salti2014shot}. These methods are designed for dense point clouds obtained by RGBD cameras, which lack generalization and robustness against noise. Deep learning has drawn much attention in recent years, and many learning-based features have been proposed. 3DMatch~\cite{zeng20173dmatch} takes 3D local patches around arbitrary interest points and extracts 3D features using a 3D convolutional neural network. PPF-net~\cite{deng2018ppfnet} utilizes PointNet~\cite{qi2017pointnet} to extract local patch features and further fuse the global context into this feature. FCGF~\cite{choy2019fully} utilizes a fully convolutional network to capture global information. It also adopts sparse convolution to efficiently extract local features in point clouds. 

These methods focus on the local feature extraction from interest points; however, interest point or keypoint detection is also important. The stable keypoints that are highly repeatable on 3D point clouds under arbitrary transformation are essential for the registration task. There is a comprehensive review of 3D keypoint detection that evaluates most handcrafted 3D keypoints~\cite{tombari2013performance}. The common trait of these methods is their reliance on local geometric information, which discards the important global context. To address these problems, USIP~\cite{li2019usip} proposes an unsupervised framework to detect keypoints. SKD~\cite{tinchev2021skd} uses saliency estimation to determine the keypoints. Some works~\cite{yew20183dfeat,bai2020d3feat} also jointly learn the keypoint detector and local descriptor.

The limitation of correspondence-based methods is the robustness of the estimator with respect to outliers and low overlaps. Then we shift from the ``3D features'' line to the ``robust estimator'' line. Several research works tried to address this problem from different perspectives. Random sample consensus (RANSAC)~\cite{fischler1981random} is a widely used robust estimator for outlier pruning. FGR~\cite{zhou2016fast} regards this problem as an optimization problem. FGR implements a Geman-McClure cost function and leverages second-order optimization to reach global registration of high accuracy. Different robust kernels are considered for point cloud registration. An elegant formulation based on Barron's kernel family~\cite{barron2019general} has been proposed in the work by Chebrolu \textit{et al.}~\cite{chebrolu2021adaptive}. TEASER by Yang \textit{et al.}~\cite{TeaserTRO2021} is the first certifiable registration algorithm that can achieve acceptable results with a large percentage of outliers. A powerful maximum clique finder~\cite{eppstein2010listing} is an important module for handling outliers in TEASER. With pruned correspondences, graduated nonconvexity~\cite{yang2020graduated} is then used for robust pose estimation. The \textit{maximum clique} problem can also be formulated as a graph-theoretic optimization problem. Parker \textit{et al.}~\cite{lusk2021clipper} present CLIPPER to solve this optimization by continuous relaxation. As for learning-based approaches, DGR~\cite{choy2020deep} proposes a differentiable scheme for closed-form pose estimation and a robust gradient-based SE(3) optimizer for refinement. PointDSC~\cite{bai2021pointdsc} utilizes a spatial-consistency guided nonlocal module for feature learning and proposes a differentiable neural spectral matching for outlier removal.

\begin{figure}[t]
\centering
\includegraphics[width=\columnwidth]{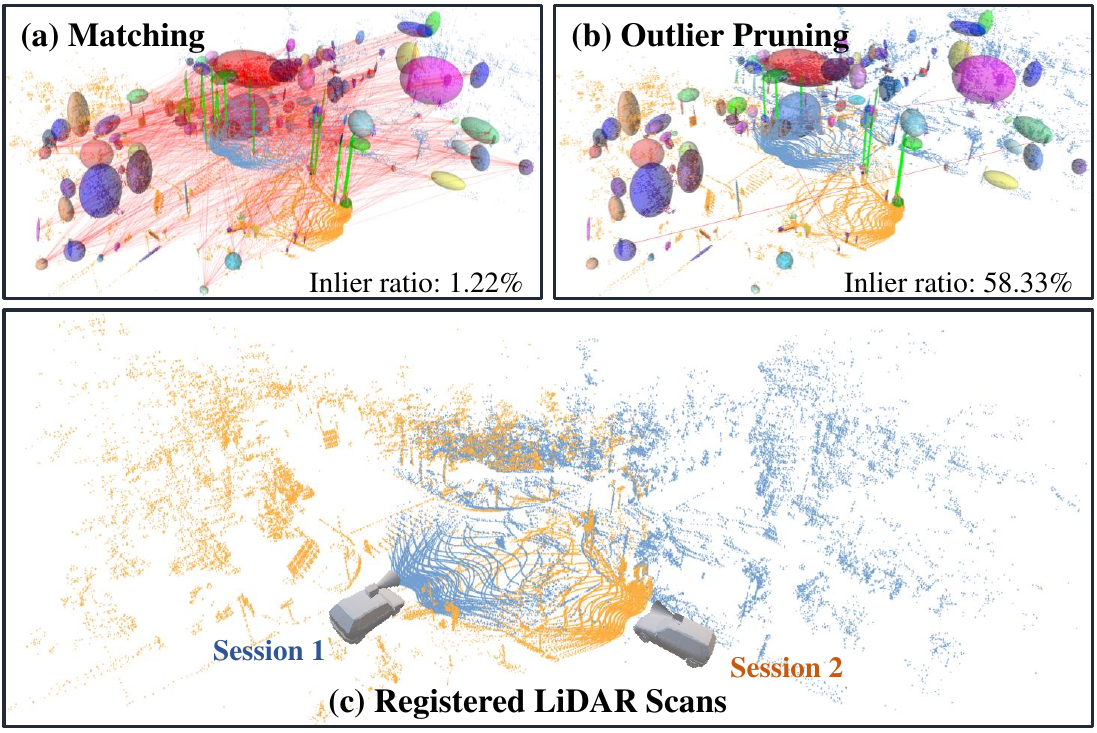}
\caption{(a) Front-end Gaussian modeling and correspondence building. (b) Back-end outlier pruning. The correspondence inlier ratio increases. (c) A challenging global registration result at a
road intersection, in which the two LiDAR point clouds are with low overlap
and a large view difference. (Source: adapted from G3Reg~\cite{qiao2023g3reg}, used with permission.) }
\label{fig:G3Reg}
\end{figure}

For global LiDAR registration, the robust estimation framework like TEASER~\cite{TeaserTRO2021} has inspired several methods recently. These works generally leverage practical considerations into the pose estimation framework. In the Quatro proposed by Lim \textit{et al.}~\cite{lim2022single}, the Atlanta world assumption is used to filter outlier correspondences in urban environments, and only one single correspondence is enough for pose estimation under the assumption. In the extension version~\cite{lim2023quatro++}, authors introduce ground segmentation into the registration framework for better LCD. G3Reg by Qiao \textit{et al.}~\cite{qiao2023g3reg} builds local Gaussians to model point cloud clusters at the front end. At the back end, G3Reg solves multiple maximum cliques for final pose estimation by considering the probability degrees of Gaussians. Figure~\ref{fig:G3Reg} shows the registration process of G3Reg.

Instead of applying an off-the-shelf robust estimator after descriptors, some works convert the entire pose estimation into the end-to-end training pipeline. Deep Closest Point (DCP)~\cite{wang2019deep} revises the original ICP pipeline to a differentiable one that can learn from data. DeepGMR~\cite{yuan2020deepgmr} is a learning-based method that leverages point-to-distribution correspondences for registration. Recently, the attention mechanism is also adopted to replace the role of feature matching and outlier filtering and thus can be used in end-to-end frameworks~\cite{predator21,shi2021keypoint,yew2022regtr}. This data-driven works~\cite {wang2019deep,yuan2020deepgmr,predator21,shi2021keypoint,yew2022regtr} are trained and validated on public point cloud datasets. More data are necessary to ensure the robustness and generalization ability needed by global localization on mobile robotics.

\subsubsection{Correspondence-free}
\label{sec:PRPE-free}

The main idea of correspondence-free methods is to register point clouds based on feature similarity. With the convergence considered, existing methods can be divided into locally convergent and globally convergent methods. The locally convergent methods stem from the optical flow in the image domain. Instead of using 3D coordinates, PointNetLK~\cite{aoki2019pointnetlk} uses PointNet~\cite{qi2017pointnet} to learn the local feature of each point and then iteratively align the learned features, which requires no costly computation of point correspondences in space. There also exist improved versions of PointNetLK framework~\cite{li2021pointnetlk,huang2020feature}. One disadvantage of this class of approaches is the iterative solver, which is sensitive to initialization and may mislead the feature learning.

Globally convergent approaches are mainly based on the idea of correlation. Like the image registration pipeline, Bulow \emph{et al.}~\cite{bulow2018scale} utilize 3D Fourier-Mellin transform to achieve globally convergent 3D registration. PHASER~\cite{bernreiter2021phaser} generates spherical frequency spectrum using  Fourier transform and Laplace fusion and registers point cloud by calculating correlation. Zhu \emph{et al.}~\cite{zhu2022correspondence} propose to learn an embedding for each point cloud in a feature space that preserves the SO(3)-equivariance property. The global convergence mostly contributed to the correlation, an inherently exhaustive search that can be evaluated effectively by spectrum decoupling. Recent work DPCN++~\cite{chen2023dpcn++} designs a differentiable phase correlation scheme with trainable networks, which could handle both 2D/3D homogeneous and heterogeneous measurements, like registering a LiDAR submap to a satellite map.

\textbf{Summary.} Point cloud registration is a popular topic but there still remain some issues for mobile robotic applications, e.g., generalization ability under an end-to-end framework and global registration with low overlap. 
In certain applications, only local registration could also provide global localization results, e.g., LiDAR LCD with local convergence-based ICP when the current pose is close to the previously stored pose. However, in such cases, global registration can provide more reliable local poses between measurements and retrieved places. Overall, if we combine approaches introduced in Section~\ref{sec:PR} and~\ref{sec:PRPE}, a complete global localization framework can be obtained in a coarse-to-fine manner: \textit{first global place recognition then followed by local pose estimation.}

Compared to place recognition-only approaches, the coarse-to-fine framework can provide precise poses for global localization tasks. The cost is that the map needs to include both global descriptors for retrieval and local metric points for state estimation. This makes the framework impracticable in large-scale environments, e.g., self-driving cars in city-scale environments for commercial use. Additionally, if place recognition fails, local pose estimation will suffer from this failure. We will introduce pose estimation-coupled place recognition to address these problems.

\subsection{Pose Estimation-coupled Place Recognition}
\label{sec:PEPR}

For the approaches explained in Section~\ref{sec:PRPE}, two separate steps are needed to handle place recognition and local pose estimation. One upgrade direction is to design a shared feature embedding or representation that place recognition and pose estimation can benefit from it. Thus, \textit{place recognition and pose estimation could share the same processing pipeline}, making the map more concise and the pipeline tighter. We name this kind of approach as pose estimation-coupled place recognition, or coupled methods for clearance in the following sections.

Note that many methods in this section use the same pre-processing techniques in Section~\ref{sec:PR}: \textit{dense points/voxel-based}, \textit{sparse segments-based and projection-based}. In this section, we mainly group these methods based on the dimension of output poses for easier understanding.

\subsubsection{3-DoF Pose Estimation}
\label{sec:PEPR-3dof}

For mobile robots working on planar surfaces, pose estimation mainly focuses on three degrees of freedom (3-DoF): position and orientation/heading (yaw angle). One of the well-known methods is scan context~\cite{kim2018scan}. The 3D point clouds are divided into azimuthal and radial bins, in which the value is assigned to the maximum height of the points in it. The similarity is the sum of cosine distances between all the column vectors at the same indexes. As the column would shift when the viewpoint of the LiDAR changes, the authors propose a rotation-invariant descriptor extracted from scan context for top-k retrieval during place recognition, then further calculate the similarity and azimuth by column shift. The rotation here is the yaw angle or the heading for mobile robots moving on planar. 

\begin{figure}[t]
\centering
\includegraphics[width=\columnwidth]{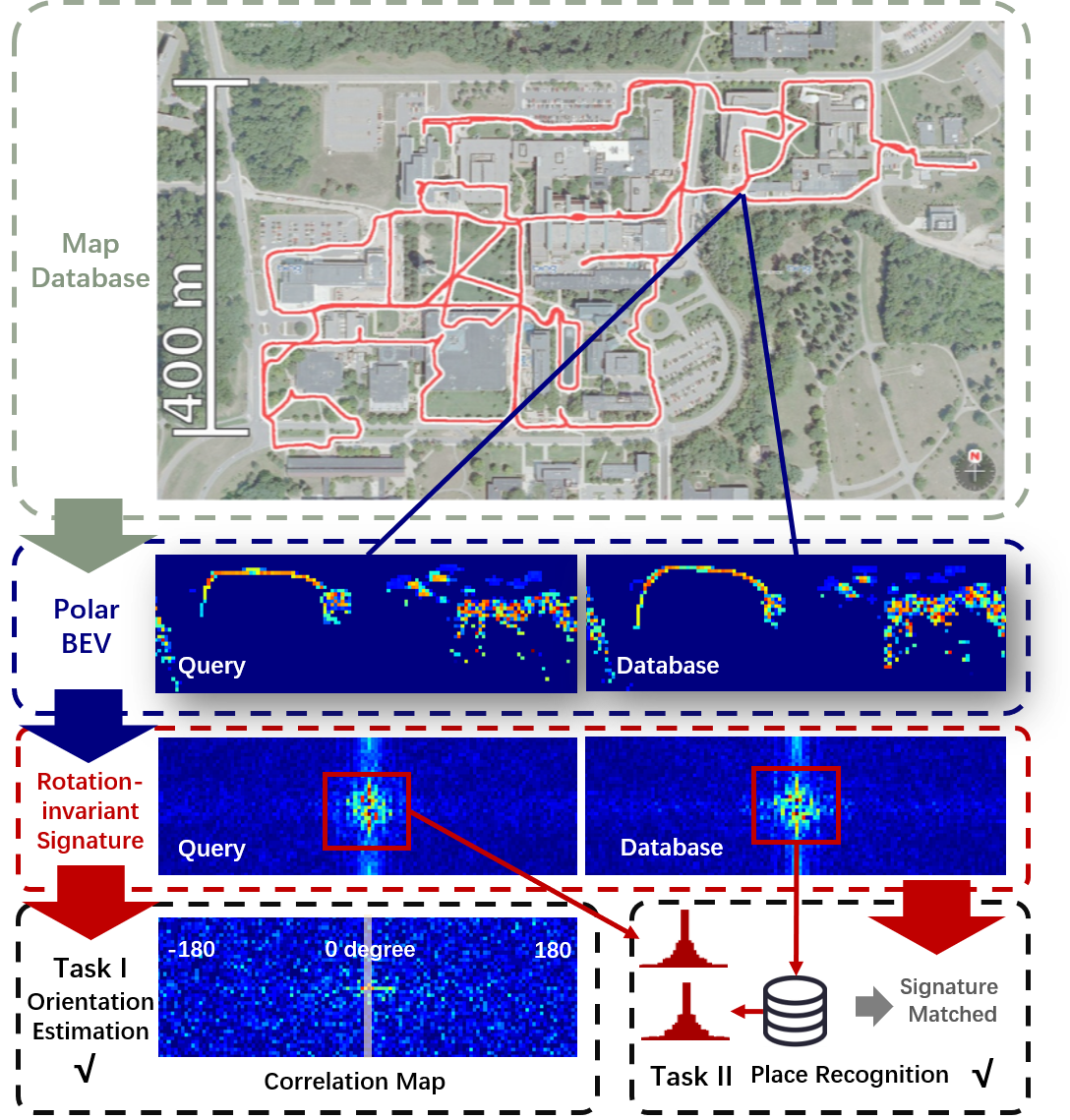}
\caption{DiSCO, differentiable scan context, jointly performs place retrieval and relative orientation estimation. The representations are polar BEV and learned rotation-invariant descriptors (Source: DiSCO~\cite{xu2021disco}, used with permission.) }
\label{fig:disco}
\end{figure}

Some following methods are designed to improve the discriminability and invariance of the original scan context~\cite{wang2020intensity,li2021ssc,SCI2019,wang2020lidar,xu2021disco,zhou2022ndd}. For example, Li \emph{et~al.}~\cite{li2021ssc} introduce the semantic labels of the point clouds. Besides feature extraction, other methods improve the efficiency of the similarity calculation process by taking advantage of the circular cross-correlation property in scan context representation. Wang \emph{et~al.}~\cite{wang2020lidar} utilize the Fourier transform to estimate the translation shift along the azimuth-related axis. Xu \emph{et~al.}~\cite{xu2021disco} propose the DiSCO, a differentiable scan context method that trains the place recognition (position) and pose estimation (orientation) in an end-to-end manner. Figure~\ref{fig:disco} shows the pipeline and representations of DiSCO for global localization.

Though scan context and its family obtain rotation-invariant descriptors for retrieval, they cannot achieve translation invariance due to the egocentric modeling process~\cite{sino22}. Few of these methods can calculate the translation drift between the query and retrieved point clouds. To relieve these limitations, scan context~\cite{kim2018scan} augments the query point cloud with root-shifted point clouds. And later in scan context++~\cite{kim2021scan}, Kim \textit{et al.} propose the Cartesian bird' s-eye view (BEV)-based descriptor for translation estimation.
RING by Lu \textit{et al.}~\cite{lu2022one} design a non-egocentric descriptor with both rotational and translational invariance. Besides place recognition and azimuth estimation, translation can also be estimated with this unified descriptor. 

Cartesian BEV image is an easily implementable choice for global LiDAR localization. Visual descriptors and matching techniques can be applied to the 2D LiDAR BEV images for 3-DoF pose estimation in~\cite{luo2021bvmatch}. Contour Context~\cite{jiang2023contour} extracts BEV-based contours that encode local information for both place recognition and pose estimation. Despite the BEV projection, the spherical projection model also transforms 3D point clouds to 2D range images for processing. OREOS~\cite{schaupp2019oreos} utilizes a convolutional neural network to extract features on the range images and generates two vectors for place recognition and azimuth estimation simultaneously. OverlapNet~\cite{chen2021overlapnet} estimates the overlap between two range images by calculating all possible differences for each pixel and calculating the azimuth taking advantage of the circular cross-correlation. An improved version, OverlapTransformer~\cite{ma2022overlaptransformer} is also proposed with a rotation-invariant representation and faster inference. The advantage of OverlapTransformer is the missing ability to provide yaw angle estimation. It is worth mentioning that the OverlapNet family uses the overlap of range images for loss function construction, which is different from the location-based loss in other learning-based methods, like contrastive loss in~\cite{yin2017efficient}, triplet and quadruplet loss in~\cite{uy2018pointnetvlad}.

\subsubsection{6-DoF Pose Estimation}
\label{sec:PEPR-6dof}


Many visual global localization frameworks extract local descriptors on images for both place recognition and the following pose estimation. Generally, the local features are aggregated into a global descriptor using methods such as bag-of-words (BoW)~\cite{galvez2012bags}, VLAD~\cite{VLAD2010}, or ASMK~\cite{asmk2013}. Meanwhile, 6-DoF poses are often extracted from visual data from using Perspective-n-Point (PnP) algorithms~\cite{Lepetit2009} based on the matched local features. Inspired by visual image matching, Shan \emph{et~al.}~\cite{shan2021robust} utilize the traditional BoW algorithm in visual place recognition for LiDAR-based global localization. Specifically, they transform the intensity of the high-resolution lidar point cloud into images and extract features based on Oriented FAST and ORB~\cite{rublee2011orb}.  The visual matching technique is also tested on~\cite{di2021visual}. However, works by Shan \textit{et al.}~\cite{shan2021robust} and Giammarino et al.~\cite{di2021visual} require a high-resolution LiDAR scanner (64 and 128 rings) to guarantee the extraction and description of local features.

Visual-inspired matching needs a projection to reduce the dimensionality of 3D point clouds. Some researchers propose to design discriminative 3D features for both global descriptor encoding and local matching, thus achieving place retrieval and 6-DoF pose estimation for global LiDAR localization. It is becoming a new research trend in the last three years (2019-2022). DH3D by Du \textit{et al.}~\cite{dh3d2020} uses flex convolution and squeeze-and-excitation block as the feature encoder and applies a saliency map for keypoint detection. Then the local features were aggregated into a global descriptor for place recognition. EgoNN by Komorowski \textit{et al}.~\cite{komorowski2021egonn} transforms the point clouds into a cylindrical occupancy map, and develops a 3D convolutional architecture based on MinkLoc3D~\cite{komorowski2021minkloc3d} for keypoint regression and description. Cattaneo \textit{et al.}~\cite{cattaneo2022lcdnet} propose an end-to-end LCDNet that can achieve both place recognition and pose estimation. LCDNet modifies PV-RCNN~\cite{shi2020pv} for local feature extraction and builds a differentiable unbalanced optimal transport~\cite{chizat2018scaling} for feature matching. BoW3D~\cite{cui2022bow3d} utilizes 3D point cloud feature LinK3D~\cite{cui2022link3d} for feature extraction and adapted BoW for global localization. Instead of point-level features, GOSMatch~\cite{zhu2020gosmatch} extracts high-level semantic objects for global localization. The authors propose a histogram-based graph descriptor and vertex descriptor taking advantage of the spatial locations of semantic objects for place recognition and local feature matching. Similarly, BoxGraph~\cite{pramatarov2022boxgraph} encodes a semantic object and its shape of a 3D point cloud into a vertex of a fully connected graph. The graph is used for both similarity measure and pose estimation. Yuan \textit{et al}.~\cite{yuan2022std} propose a novel triangle-based global descriptor, stable triangle descriptor (STD) for place recognition and relative pose estimation. STD keeps a hash table as the global descriptor and place recognition is achieved by voting of triangles in the table.

All the methods above achieve place recognition by nearest neighbor search or exhaustive comparisons on global descriptors. Several works only use local keypoints or features to build coupled methods, and there are no global descriptors for place retrieval. Bosse and Zlot~\cite{bosse2009keypoint,bosse2013place} extract and describe keypoints for both place candidate voting and 6-DoF pose estimation. Inspired by the work of Bosse and Zlot~\cite{bosse2013place}, Guo \textit{et al.}~\cite{guo2019local} design an intensity-integrated keypoint and also propose a probabilistic voting strategy. Steder \textit{et al.}~\cite{steder2010robust} propose to match point features on range images and score potential transformations for final pose estimation. Instead of extracting features on point clouds, Millane \textit{et al.}~\cite{millane2019free} introduce a SIFT-inspired~\cite{lowe1999object} local feature based on the distance function map of 2D LiDAR submaps. Experiments validate that using free space for submap matching performs better compared with using occupied grids. 

\textbf{Summary.} Pose estimation-coupled place recognition provides not only the 
 retrieved place but also a 3-DoF (or with only a 1-DoF yaw angle) or a full 6-DoF pose. Common evaluation metrics include both precision-recall for retrieval and quantitative errors compared to ground truth orientation or position.

One might ask about the advantages of using such methods compared to the previous two-step pipeline using place retrieval (Section~\ref{sec:PR}) followed by precise pose estimation (Section~\ref{sec:PRPE}). The potential advantages are three folds:
\begin{itemize}
\item Lightweight map. Dense point maps limit mobile robotic applications in large-scale environments, especially for resource-constrained vehicles. If place recognition and pose estimation share the same feature or representation, fewer data and sparser keyframes could support global localization in such conditions, making the entire map more lightweight to use.
\item Geometric verification. For place recognition methods, a key issue is to verify whether a retrieved place is correct. Pose estimation results can be used as geometric verification to filter incorrect places. This filtering strategy has been applied in several coupled methods~\cite{zhu2020gosmatch,yuan2022std}.
\item Initial guess for refinement. If an accurate pose is required, a local point cloud registration (Section~\ref{sec:PRPE}) is necessary for pose refinement. Coupled approaches can provide an initial guess for such local registration modules, thus improving the accuracy and efficiency of pose estimation. As reported in LCDNet~\cite{cattaneo2022lcdnet}, the initial guess by LCDNet significantly reduces runtime and metric errors when applying ICP registration.
\end{itemize}
Overall, the pose estimation and place recognition are coupled in this subsection, but keyframes or places are still needed in a pre-built map database. The one-stage approaches will be introduced in the following subsection, which only requires a global map for global localization.

\subsection{One-stage Global Pose Estimation}
\label{sec:PE}

The two-stage methods using place recognition and pose estimation techniques have shown successful operations in various datasets and applications. Thus a natural question is raised: can we achieve global localization by directly matching on a global map without separating the place? The answer is yes and some approaches can achieve one-stage global pose estimation. The majority of these approaches can be classified into two categories based on how to estimate the pose: in a traditional closed form or in an end-to-end manner.

\subsubsection{Feature-based Matching}
\label{sec:PE-feature}

\begin{figure}[t]
\centering
\includegraphics[width=\columnwidth]{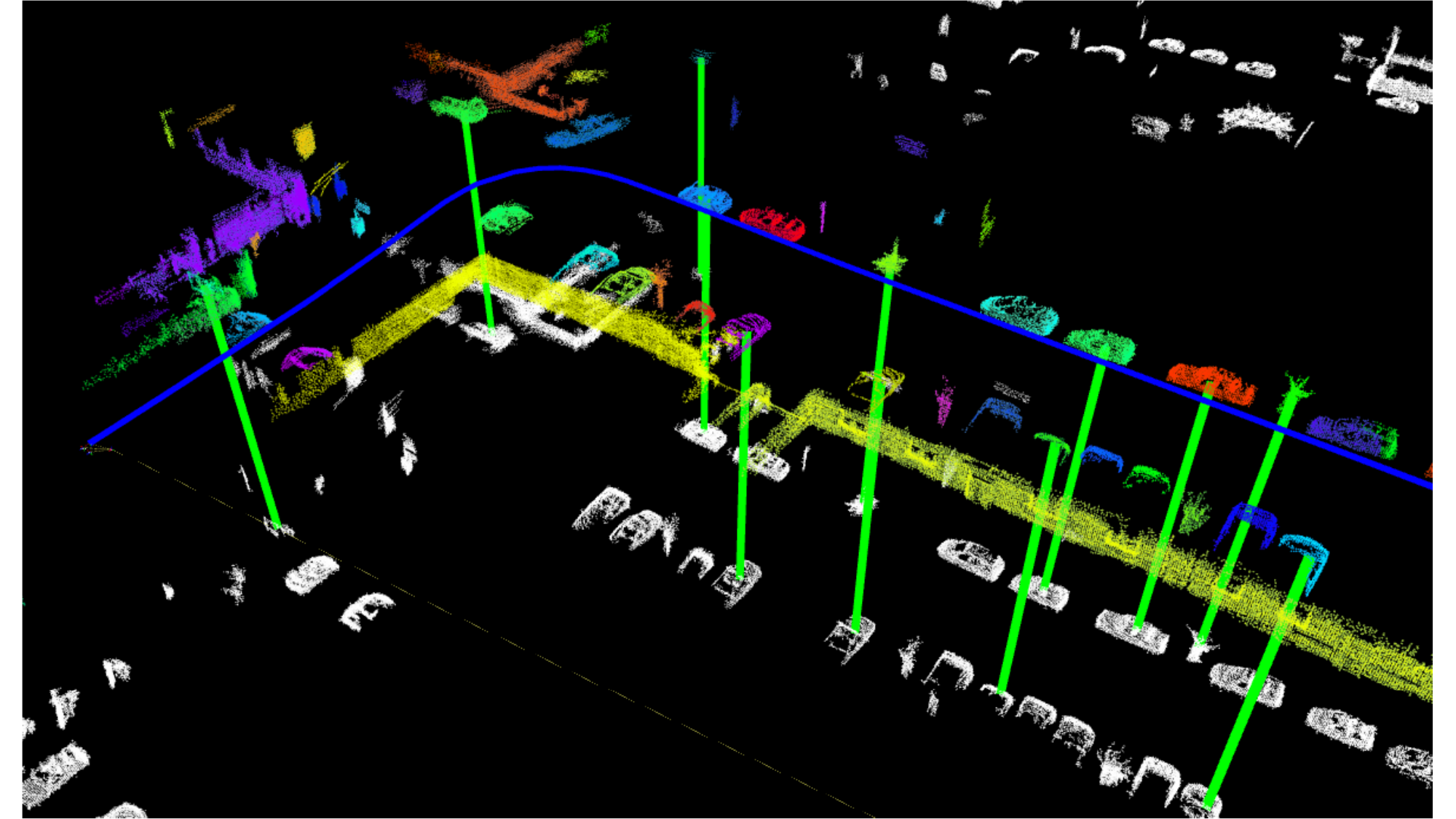}
\caption{SegMatch achieves global localization at the level of segment features rather than conventional keypoints. Different colors represent the segmentation results. Segment matches are indicated with green lines. (Source: SegMatch~\cite{dube2017segmatch}, used with permission.) }
\label{fig:segmatch}
\end{figure}

One representative work is SegMatch proposed by Dub{\'e} \emph{et~al.}~\cite{dube2017segmatch} in 2017, with results shown in Figure~\ref{fig:segmatch}. SegMatch first segments dense LiDAR map points to clusters with ground removal and then extracts features based on eigenvalues and shapes of segments. Random forest classifier is trained and applied to boost feature matching. Finally, matched candidates are fed into RANSAC for 6-DoF pose estimation. The handcrafted descriptors were extended to data-driven SegMap~\cite{dube2018segmap} with the help of deep neural networks. In~\cite{cramariuc2021semsegmap}, SemSegMap is proposed by integrating visual information into point cloud segmentation and feature extraction. SegMatch and its ``family members'' are validated in urban and disaster environments.

Inspired by SegMatch scheme, Tinchev \emph{et~al.}~\cite{tinchev2018seeing} propose Natural Segmentation
and Matching (NSM)  for global localization in a more natural environment. The insight is a novel hybrid descriptor and is more robust to different points of view than the baseline SegMatch. Similarly, NSM has also been extended to a deep learning version in~\cite{tinchev2019learning}. In these feature-based global matching methods~\cite{dube2017segmatch,dube2018segmap,cramariuc2021semsegmap,tinchev2018seeing,tinchev2019learning}, point cloud segments are aligned with low-dimensional and distinctive descriptors for global matching. The global segment-based maps can support not only loop closing for consistent mapping but also pose tracking for online localization.

In addition to the SegMatch and NSM schemes, recent works also propose to achieve one-shot global localization with semantic objects. Ankenbauer~\cite{ankenbauer2023global} \textit{et al.} design a re-localization scheme that leverages the graph-theoretic knowledge. The scheme could register the observed semantic objects to the prior object map and is validated on both a planetary rover dataset and the KITTI dataset. The graph-theoretic matching is also leveraged in the single-shot method proposed by Matsuzaki \textit{et al.}~\cite{matsuzaki2023single}. First, semantic information bridges the modality gap between visual images and LiDAR maps with correspondences; and then outliter pruning is achieved by maximum clique solver for robust pose estimation.

Generally, these segment matching-based approaches rely on the segmentation results, as mentioned in Section~\ref{sec:PR-segments}. A robot traveling is a good choice to accumulate dense 3D point clouds first, and then segmentation and segment matching are performed. Hence, these methods are less efficient compared to two-stage approaches for some specific tasks, like fast re-localization with sparse scans. Additionally, the reliance on segments could make these methods easily fail in challenging scenes, e.g., in a featureless flat field or a man-made environment with too many repetitive structures.

\subsubsection{Deep Regression}
\label{sec:PE-deep}

With the popularity of deep learning, several researchers propose to regress global robot pose directly in an end-to-end fashion, just like PoseNet~\cite{kendall2015posenet} for visual re-localization. 
Similarly, Wang \emph{et~al.}~\cite{wang2021pointloc} propose a learning-based PointLoc for LiDAR global pose estimation. The backbone is an attention-aided PointNet-style architecture~\cite{qi2017pointnet} for 6-DoF pose regression. This end-to-end manner is completely data-driven without conventional pose estimation processing. Lee \emph{et~al.} ~\cite{lee2022learning} convert the global localization as an unbalanced point registration problem, and propose a hierarchical framework UPPNet to solve this problem. Specifically, UPPNet first searches the potential subregion in a large point map and then achieves pose estimation via local feature matching in this subregion. UPPNet can also be trained in an end-to-end fashion.

\textbf{Summary.} Feature-based one-stage approaches do not use discrete places or locations for place recognition. They are suitable for loop closure detection in a small area, i.e., the solution space $\left| \mathbb{X} \right|$ is reduced to a smaller size in Equation~\ref{eq:sing-shot}. However, the downside is that it is challenging to re-localize a robot from scratch using partial local features in a large feature map.

As for one-stage deep regression approaches, though LiDAR scanner provides rich structural information, the metric estimation is not competitive~\cite{wang2021pointloc,lee2022learning} compared to conventional two-stage methods in Section \ref{sec:PRPE} and \ref{sec:PEPR}. We consider one-stage pose regression is a promising research direction in the era of big data, but still remain many issues to solve, e.g., how to improve the interpretability and generalization ability of these end-to-end methods.

\section{Sequential Global Localization}
\label{sec:sequential}

Section~\ref{sec:single} reviews related single-shot global localization approaches that take a single LiDAR point cloud as the input. As previously analyzed in Section~\ref{sec:problem}, the map size $\left| \mathbf{M} \right|$ is generally much larger than the size of single point cloud $\left| \mathbf{z}_t \right|$, while the single-shot global localization methods can not guarantee the localization success in challenging scenes. On the other hand, LiDAR sensor provides high-frequency point measurements, and sequential point clouds can be obtained when the robot travels a distance. Thus taking multiple measurements $\mathbf{Z}_{t} \triangleq \left\{ \mathbf{z}_{k=1}, \cdots \mathbf{z}_t\right\}$ could enhance the global localization performance choice with robot moving. This section reviews methods that use sequential LiDAR inputs for global pose estimation. Note that place retrieval methods in Section~\ref{sec:single} can be integrated as a front-end matching in frameworks of this section.

\begin{figure}[t]
\centering
\subfigure[Sequential place matching]{
	\includegraphics[width=\columnwidth]{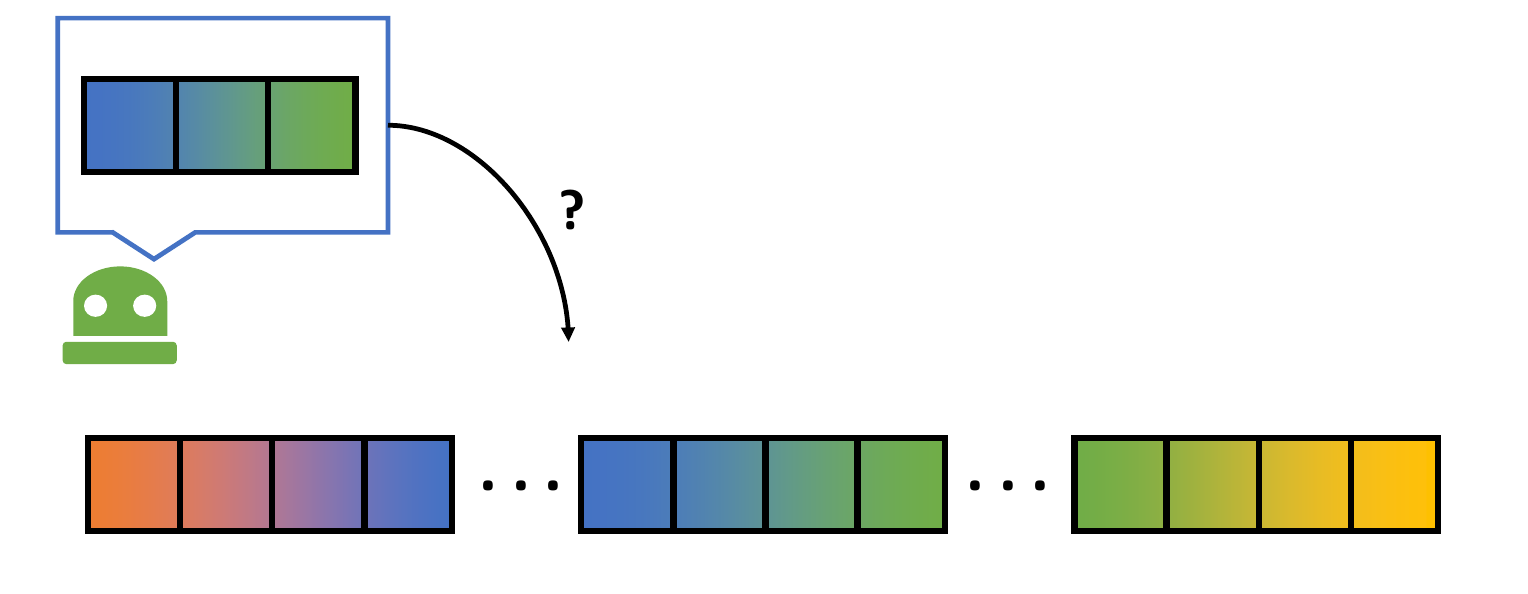}}
\subfigure[Sequential metric localization]{
	\includegraphics[width=\columnwidth]{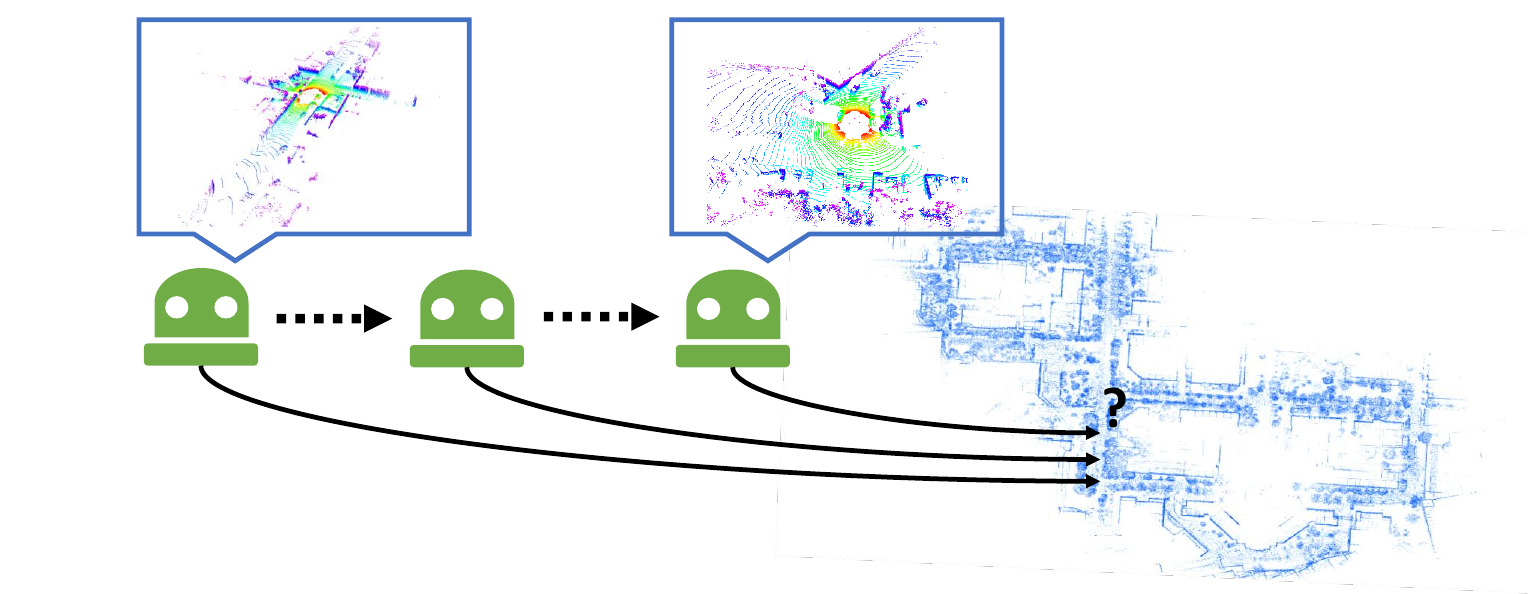}}
\caption{Graphical illustration of two categories of sequential global localization using LiDAR. Sequential place matching fuses the multiple single-shot global localization results using sequence information. Sequential metric localization fuses the multiple non-global measurements thus a back-end is required for multi-hypotheses filtering. The maps are also different for these two types of approaches.
}
\label{fig:seq-global}
\end{figure}

Sequential global localization can be divided into two categories determined by its map and the use of place recognition and pose estimation. One is \textit{sequential place matching} on keyframe-based submaps and the other is \textit{sequential metric global localization} on metric maps. The former provides a retrieved place as a localization result and is performed on keyframe-based submaps. The latter estimates an accurate pose on a metric map and is generally based on a global metric map. In addition, sequential metric localization generally requires a state estimator at the back end to track non-global localization results. A graphical illustration is presented in Figure~\ref{fig:seq-global}.

As analyzed in Section~\ref{sec:problem}, we consider that sequential-based approaches can also be classified into two categories: \emph{batch processing} and \emph{recursive filtering}. The difference is how to handle sequential information for global pose estimation: batch methods handle a batch of information to estimate the entire robot trajectory via retrieval or optimization (Equation~\ref{eq:sequential-batch}); filtering methods estimate the pose under Bayesian filtering or similar techniques (Equation~\ref{eq:sequential-markov}). We will also talk about this taxonomy as an underlying theme in the following two subsections. We present several representative works in Table~\ref{tab:sequential-global}.


{
\renewcommand{\arraystretch}{1.5} 
\begin{table*}
\centering
\caption{\textbf{Representative Studies of Sequential Global LiDAR Localization}}
\label{tab:sequential-global}
\begin{tabularx}{\textwidth}{l l l l X }
\hline
Name & Pose  & Handling & Backbone \\
\hline
SeqLPD \cite{liu2019seqlpd} & Place  &  Batch Processing & LPDNet~\cite{liu2019lpd} + Sequence Matching \\
Yin \textit{et al.} \cite{yin2022fast} & Place  & Recursive Filtering & SphereVLAD + Hierarchical Particle Filter \\
SeqOT \cite{ma2022seqot} & Place  & Batch Processing & Multi-scan Transformer + GeM Pooling \\
\hline
Dellaert \textit{et al.} \cite{dellaert1999monte} & 3-DoF  & Recursive Filtering & Monte Carlo Localization (MCL) \\
Jonschkowski \textit{et al.}  \cite{jonschkowski2018differentiable} & 3-DoF  & Recursive Filtering & Differentiable Particle Filter (Differentiable MCL) \\
Chen \textit{et al.} \cite{chen2021deep} & 3-DoF  & Recursive Filtering &  Deep Samplable Observation Model + Adaptive
Mixture MCL \\
Gao \textit{et al.} \cite{gao2019novel} & 3-DoF  & Recursive Filtering & Feature Matching + Odometry + Multiple Hypothesis Tracking \\
GLFP \cite{wang2019glfp} & 6-DoF  & Batch Processing & Landmark Association + Odometry + Factor Graph \\
\hline
\end{tabularx}
\end{table*}
}

\subsection{Sequential Place Matching}
\label{sec:seqMatch}

Probabilistic or sequential matching can help improve the visual localization success rate, which has been validated in several classical visual systems: FAB-MAP~\cite{cummins2008fab}, SeqSLAM~\cite{milford2012seqslam,milford2015sequence}, the work of Naseer \textit{et al.}~\cite{naseer2018robust}, and Vysotska and Stachniss~\cite{vysotska2019effective}. FAB-MAP first builds an appearance-based BoW for single image retrieval and then formulates recursive Bayesian filtering for global localization. An extended version FAB-MAP 3D~\cite{paul2010fab} also models spatial information to improve the robustness of the framework. The filtering technique of FAB-MAP family could handle sequential measurements, but will easily crash when single-shot place recognition fails in challenging scenes. In SeqSLAM~\cite{milford2012seqslam}, a sequence-to-sequence matching strategy is proposed to find the location candidates in an image similarity matrix. SeqSLAM processes a batch of images compared to filtering-based methods, making the whole system more robust. The SeqSLAM family has demonstrated its success in both handcrafted features~\cite{milford2012seqslam} and data-driven features~\cite{milford2015sequence}. Naseer \textit{et al.}~\cite{naseer2018robust} propose to use a network flow to handle batch image matching and maintain multiple route hypotheses in parallel. Global visual matching method is also proposed in~\cite{vysotska2019effective} for re-localization, in which the map database contains multiple sequences for graph-based search.

The LiDAR-based sequential matching has been inspired by visual methods in recent years. Liu \emph{et~al.}~\cite{liu2019seqlpd} propose to use LPD-Net~\cite{liu2019lpd} for front-end place recognition, and design a coarse-to-fine sequence matching strategy for global localization. The designed strategy improves the place retrieval performance compared with single-shot LPD-Net. Yin \emph{et~al.}~\cite{yin2022fast} present a particle-aided fast matching scheme in large-scale environments based on sequential place recognition results, which is generated by SphereVLAD in Section~\ref{sec:PR-proj}. From the viewpoint of state estimation,~\cite{liu2019seqlpd} handles batch information while~\cite{yin2022fast} recursively estimates the locations. Recent work SeqOT~\cite{ma2022seqot} generates one global descriptor for a sequence of range images, rather than multiple descriptors in its previous version~\cite{ma2022overlaptransformer}. Specifically, a novel end-to-end transformer is built to handle spatial and temporal information fusion.

All these methods above, visual- or LiDAR-based, aim at estimating the most likely (highest probability) match on topological keyframe-based submaps (Section~\ref{sec:KM}). The evaluation of these methods is the same with place recognition-only approaches in Section~\ref{sec:PR}.

\subsection{Sequential Metric Localization}
\label{sec:seqMetric}

If the map has a geometric representation, like occupancy grids and landmarks, it can enable metric pose estimation for mobile robots, making sequential global localization more practical. 

Particle filter localization, also known as sequential Monte Carlo Localization (MCL) in the robotics community, is a widely used recursive state estimation back-end~\cite{dellaert1999monte}. Unlike the Kalman filters family, MCL is non-parametric Bayesian filtering without assuming the distributions of robot states. More specifically, it uses a group of samples to represent the robot state, which is naturally suitable for global localization tasks especially when the robot pose has a multi-modal distribution. Researchers have proposed multiple extended versions to improve the robustness and efficiency of the original MCL. Maintaining a large set of particles is computationally expensive. Adaptive MCL~\cite{fox2001kld} can sample particles in an adaptive manner using the Kullback–Leibler divergence. In~\cite{stachniss2005mobile}, segmented patch maps are integrated into MCL framework, making it applicable in indoor non-static environments. Most LiDAR sensors can also intensity information as reflection properties of surfaces. Bennewitz \textit{et al.}~\cite{bennewitz2009utilizing} use these reflection properties to improve the observation model of MCL, and it achieves faster convergence for re-localization. Recent work~\cite{zimmerman2022robust} also integrates human-readable text information into MCL for localization, making it more robust to structural changes in buildings. Due to its simplicity and effectiveness, MCL is also used in various low-dimensional navigation tasks beyond global localization, such as robotic pose tracking~\cite{yin2022radar} and exploration tasks~\cite{stachniss2005information}.

Currently, MCL is one of the gold standards in multiple robot navigation toolkits~\cite{montemerlo2003perspectives,zheng2021ros}. Indoor LiDAR MCL is well studied and has been widely deployed for commercial use, e.g., applying to a home cleaning robot. A recent trend of indoor LiDAR localization is to use building architectures as maps, e.g., structural computer-aided design (CAD)~\cite{boniardi2017robust,zimmerman2023ral} and semantic building information modeling (BIM)~\cite{yin2023semantic,hendrikx2021connecting}. These maps are easy to obtain and keep sparse but critical information of environments, like walls and columns. The use of such maps makes the localization free of pre-mapping for long-term operations. MCL can also be used for localization on floor plan maps~\cite{boniardi2017robust,zimmerman2023ral}. Experiments show that such cheap maps can also support indoor robot localization.

Modern MCL methods integrate discrete place recognition techniques into the filtering framework, making MCL applicable in large-scale outdoor environments. Yin \emph{et~al.}~\cite{yin2018locnet} propose to use the Gaussian mixture model to fuse multiple place recognition results, and then integrate it into the MCL system as a measurement model. With the convergence of MCL, a coarse pose can be generated as an initial guess for accurate ICP refinement. The observability of orientation is also proofed in its extended version~\cite{yin20193d}. Similarly, Chen \emph{et~al.}~\cite{chen2020iros} use their OverlapNet~\cite{chen2021overlapnet} to extract features of submaps in a global map and propose a new observation model for MCL by comparing the similarity between the current feature and the stored features to achieve global LiDAR localization. Sun \emph{et~al.}~\cite{sun2020localising} and Akai \emph{et~al.}~\cite{akai2020hybrid}  propose to fuse deep pose regression and MCL to build a hybrid global localization, in which deep pose regression could provide a 3-DoF or 6-DoF from an end-to-end neural network. The methods above~\cite{yin20193d,chen2021overlapnet,sun2020localising,akai2020hybrid} typically discrete the pose and map space for fast convergence of MCL in large-scale environments. A deep learning-aided samplable observation model was proposed in~\cite{chen2021deep}, named DSOM. Given a 2D laser scan and a global indoor map, DSOM can provide a probability distribution for MCL on the global map, thus making particle sampling focus on high-likelihood regions.

The advancements caused by deep learning methods also affect the back-end state estimator of the MCL system. Jonschkowski \emph{et~al.}~\cite{jonschkowski2018differentiable} design a differentiable particle filter (DPF) scheme for robot pose tracking and global localization. The whole DPF pipeline includes differentiable motion and measurement models, and a belief update model for particles, making the DPF trainable in an end-to-end manner. In the Differentiable SLAM-net proposed by Karkus \textit{et al.}~\cite{karkus2021differentiable}, DPF is encoded into a trainable visual SLAM for indoor localization. LiDAR-based particle filter is quite mature and there is no LiDAR-based DPF currently. But we consider differentiable state estimator could be a promising direction in this era of big data.

We also notice that there exist other frameworks that can achieve sequential metric global localization. Multiple hypotheses tracking (MHT)~\cite{thrun2002probabilistic} is a possible solution to the global localization problem. An improved MHT framework is proposed in~\cite{gao2019novel}, and authors design a new structural unit encoding scheme to weight hypotheses. Hendrikx \textit{et al.}~\cite{hendrikx2022local} propose to build a hypotheses tree for indoor global localization. A global feature map is required for this method and explicit data associations are used to check the hypotheses. Wang \emph{et~al.}~\cite{wang2019glfp} present a factor graph-based global localization from a floor plan map (GLFP). GLFP integrates odometry information and landmark matching into a factor graph when the robot travels. Compared to the filtering family (MCL and MHT), GLFP handles a batch of information for global pose estimation, which is similar to SeqSLAM in Section~\ref{sec:seqMatch}. The landmark matching in~\cite{wang2019glfp} provides global position information for factor graph optimization. In the works of Wilbers \textit{et al.}~\cite{wilbers2019approximating}, researchers employ graph-based sliding window approaches to fuse outdoor landmark matching and odometry information. Some other sensor information could also provide global positions for mobile robots. Merfels and Stachniss \textit{et al.}~\cite{merfels2016pose} fuse global poses from GNSS and odometry information to achieve self-localization for autonomous driving. Lastly, in this subsection, the evaluation typically contains metric pose estimation on the map, e.g., using Root Mean Square Error (RMSE), which differs from approaches in Section~\ref{sec:seqMatch}.

{
\renewcommand{\arraystretch}{1.5} 
\begin{table*}
\centering
\caption{\textbf{Representative Studies of Cross-robot Localization}}
\label{tab:cross-robot}
\begin{tabularx}{\textwidth}{l l l X }
\hline
Name & Environment & Robot Team &  Backbone and Highlight \\
\hline
DARE-SLAM \cite{ebadi2021dare} & Subterranean & Heterogeneous  &  LCD by Feature Matching + Degeneracy-aware LiDAR SLAM \\
LAMP 2.0 \cite{chang2022lamp} & Subterranean & Heterogeneous  &  LCD by GICP + Outlier-robust PGO  \\
DiSCo-SLAM \cite{huang2021disco} & Park & Wheeled & LCD by Scan Context~\cite{kim2018scan} + PCM + PGO \\
DCL-SLAM \cite{zhong2022dcl} & Campus & Wheeled &  LCD by LiDAR Iris~\cite{wang2020lidar} + PCM + PGO \\
RING++ \cite{xu2022ring++} & Campus & Legged & LCD by roto-invariant RING~\cite{lu2022one} + ICP + PGO \\
\hline
\end{tabularx}
\footnotesize  \textbf{PGO: Pose Graph Optimization, PCM: Pairwise Consistent Measurement Set Maximization }
\end{table*}
}

\section{LiDAR-aided Cross-robot Localization}
\label{sec:cross}

The review in Section~\ref{sec:single} and Section~\ref{sec:sequential} mainly focuses on single robot-based global LiDAR localization. Global localization can also be deployed into multi-robot systems for cross-robot localization, which is a new trend in the robotics community. More concretely, one robot performs mapping and another robot globally estimates its pose on this map, and vice versa. 

\subsection{LiDAR-aided Multi-robot System}
\label{sec:MRS}

In practice, the multi-robot system is a broad topic that involves many subproblems that are not the main concerns of this paper, such as communication bandwidth and computation efficiency. As for system architecture, both \textit{distributed} multi-robot systems and \textit{centralized} servers~\cite{bernreiter2022collaborative,cramariuc2022maplab} could work well in different scenarios. We also note that customized scan matching is proposed for point cloud map fusion and collaborative robots~\cite{yue2022aerial}. These localization methods are based on offline map appearance, while this section mainly focuses on incremental keyframe-based cross-robot localization. Several representative works are listed in Table~\ref{tab:cross-robot}.

\begin{figure*}[t]
\centering
\subfigure[LiDAR map and wheeled robots team~\cite{zhong2022dcl}]{
	\includegraphics[width=8.5cm]{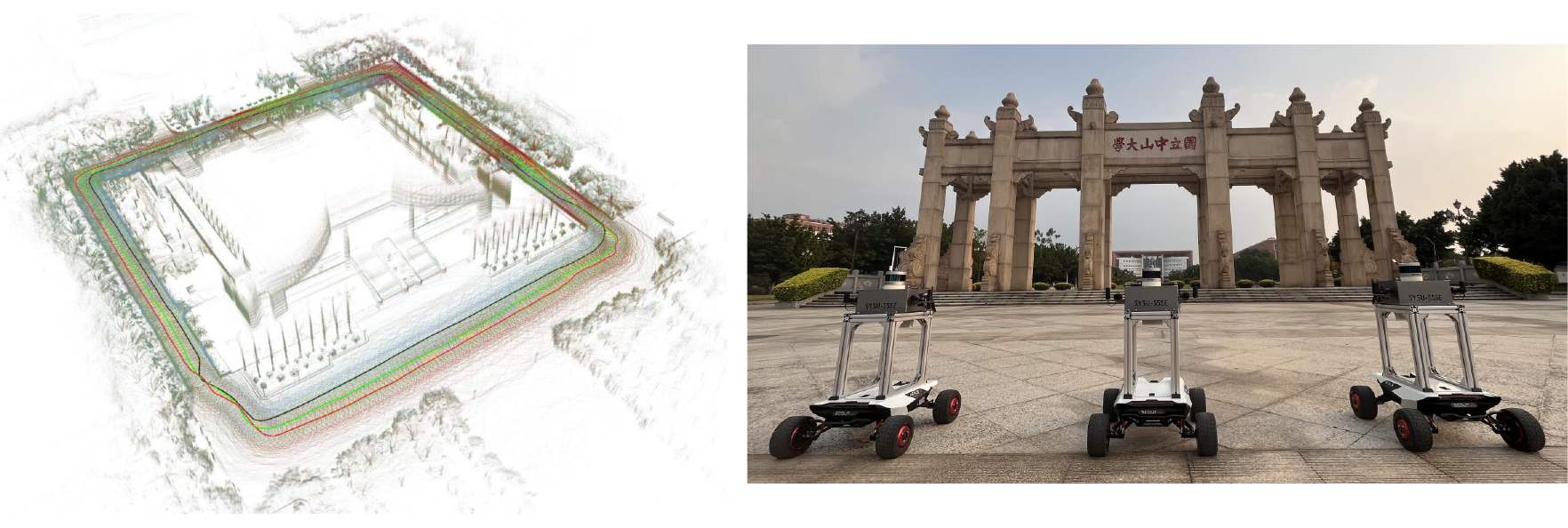}}
\subfigure[LiDAR map and legged robots team~\cite{xu2022ring++}]{
	\includegraphics[width=8.5cm]{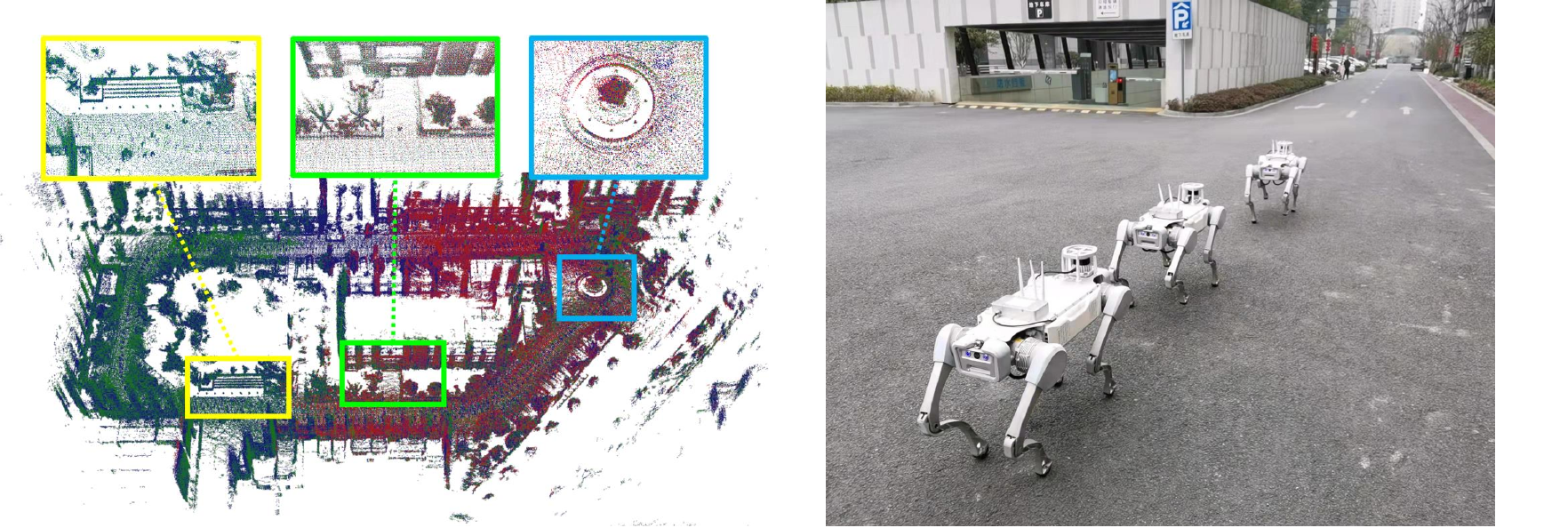}}
\caption{ Qualitative results from DCL SLAM~\cite{zhong2022dcl} and RING++~\cite{xu2022ring++} (used with permission). Different colors represent LiDAR maps and trajectories generated from different mobile robots: wheeled robots in DCL SLAM and legged robots in RING++.
}
\label{fig:cross-robot}
\end{figure*}

Over the last two decades, there has been a growing demand for autonomous exploration and mapping of various environments, ranging from outdoor cluttered and underground environments to complex cave networks. Due to this, multi-robot SLAM, a critical solution for navigation in GNSS-denied areas where prior maps are unavailable, is receiving more attention. The recent DARPA Subterranean (SubT) Challenge, a three-year global competition that ended in 2021, aimed to demonstrate and advance the state-of-the-art in mapping, localization, and exploration of complex underground settings and has been particularly important in improving multi-robot SLAM. The multi-robot SLAM architectures adopted by the six SubT teams are summarized in the survey by Ebadi \textit{et al}.~\cite{ebadi2022present}. Although many loop closure methods were proposed then, most of the teams detected loop closure candidates simply by calculating the distance between the current keyframe and another keyframe in the factor graph. This LCD strategy is the same as DARE-SLAM~\cite{ebadi2021dare} and LAMP~\cite{chang2022lamp,denniston2022loop}. The simple but effective LCD used in these methods mainly relies on the setting that all robots start to move in the same starting region. In this context, with a high-precision LiDAR odometry~\cite{zhang2014loam,zhao2021super} at the front end, the distance-based LCD can work well in a relatively small (\textless 5 km) area. There is no need to customize a complex place retrieval module in the robotic system~\cite{ebadi2022present}.

Despite the underground exploration, robots might not always be able to begin a task at the same location, as in large-scale search and rescue tasks. As a result, place recognition that does not rely solely on initials is necessary. In an aerial-ground collaborative manner, He \emph{et~al.}~\cite{he2020ground} extract obstacle outlines from submap point clouds and generate thumbnail images. The thumbnail images are converted into compact place descriptors by applying NetVLAD~\cite{arandjelovic2016netvlad}. DiSCo-SLAM~\cite{huang2021disco} firstly adopts LiDAR-based global descriptor, scan context~\cite{kim2018scan}, to perform place recognition in a distributed manner. The lightweight scan context descriptor makes the real-time application possible, although there are no field experiments with multi-robots in this paper. RDC-SLAM~\cite{xie2021rdc} utilizes a place-recognition-only global descriptor, called DELIGHT~\cite{cop2018delight}, to reduce time consumption. In the relative pose estimation part, eigenvalue-based segment descriptors are proposed to achieve feature matching. DCL-SLAM~\cite{zhong2022dcl} assesses the performance of LiDAR-Iris~\cite{LidarIris}, M2DP~\cite{he2016m2dp} and scan context~\cite{kim2018scan} and finally uses the effective and rotation-invariant LiDAR-Iris for loop closure detection. Unlike the system-oriented research mentioned above, RING++~\cite{xu2022ring++} presents a general non-learning framework to achieve roto-translation invariance with various local features while estimating the relative 3-DoF pose. The roto-translation invariant property and the robust pose estimator allow the multi-robot system to sample places along a long distance while being computationally and memory efficient. Figure~\ref{fig:cross-robot} shows the real-world experimental results using DCL SLAM~\cite{zhong2022dcl} and RING++~\cite{xu2022ring++}.

\subsection{Cross-robot Back-end}
\label{sec:CRBE}

From the systems above~\cite{he2020ground,huang2021disco,xie2021rdc,zhong2022dcl,xu2022ring++}, it can be concluded that inter-robot LCD methods enable cross-robot localization in large-scale environments. However, no LiDAR LCD method can provide perfect loop closures without false positives. The false positives are outliers that make estimation systems unstable and inaccurate. More specifically, almost all cross-robot localization systems are built on graph optimization frameworks~\cite{kummerle2011g,dellaert2012factor}. These false positives provide inconsistent links between pose nodes. The optimization may not converge to a correct solution in such conditions. This problem exists not only in LiDAR-aided cross-localization but also in other SLAM-related problems with different sensors.

There exist mainly two ways to handle this problem: one is to build an outlier rejection module or similar techniques \textit{before} the optimization; the other is to design robust kernels or functions that could reduce the impacts of outliers \textit{during} the graph optimization. These two types of approaches aim at improving the robustness of graph optimization with inconsistent edges generated from LCD methods. Outlier rejection modules are independent of back-end estimators. RANSAC~\cite{fischler1981random} is a popular method for outlier rejection, which iteratively estimates a model from sampled data (loop closures). Olson \textit{et al.}~\cite{olson2005single} propose a graph theory-based outlier rejection method, named single-cluster graph partitioning (SCGP). Note that the graph is generated from the adjacency matrix of pose nodes and not the pose graph for SLAM. SCGP estimates a pairwise consistency set as the final result via clustering in this graph. SCGP shows competitive performance with RANSAC. Similarly, the outlier rejection is formulated as a maximum set estimation problem in the work by Carlone \textit{et al.}~\cite{carlone2014selecting}. 

In 2018, Enqvist \textit{et al.} present a pairwise consistency maximization (PCM)~\cite{mangelson2018pairwise} for consistent mapping with loop closures. PCM first builds a binary consistency graph by checking all loop closures of each other. The criteria of consistency check are formulated based on the transformation from odometry modules and loop closures. PCM aims at estimating the maximum pairwise internally consistent set, which is the maximum clique problem in graph theory (the same problem mentioned in TEASER~\cite{TeaserTRO2021}, see Section~\ref{sec:PRPE}). A fast clique solver by Pattabiraman \textit{et al.}~\cite{pattabiraman2015fast} is adopted in PCM~\cite{mangelson2018pairwise}. PCM has been validated in aforementioned LiDAR-aided cross-robot localization systems~\cite{huang2021disco,zhong2022dcl}, and other robot localization and mapping systems in recent years~\cite{xu2022omni,tian2022kimera}. 



Despite the outlier rejection, the other way is to reduce the impacts of false loops during the pose graph optimization. S{\"u}nderhauf and Protzel~\cite{sunderhauf2012switchable} formulate switchable constraints in the optimization. The added constraints follow different outlier rejection policies, and they can turn on or off loop closures. An iterative approach RRR is designed in~\cite{latif2013robust} and it identifies true positives via clustering consistent loop closures. These robust functions are typically integrated into pose graph optimization frameworks to improve the robustness~\cite{kummerle2011g,dellaert2012factor}. In addition to identifying loop edges, researchers also design and use robust kernels in the optimization, such as Gemen-McClure kernel~\cite{zhang1997parameter}, Huber kernel~\cite{zhang1997parameter}, and Max-Mixture kernel~\cite{olson2013inference}. In 2023, McGann \textit{et al.}~\cite{mcgann2023robust} present the riSAM which leverages graduated nonconvexity techniques~\cite{yang2020graduated} in the pose graph optimization. riSAM achieves better performances compared with other robust estimation techniques. In real-world applications, we consider a combination use of outlier rejection and robust estimator could be a practical choice to handle false positive loop closures.

In summary, cross-robot localization is becoming a promising direction for future study. It involves multiple topics of robotics, such as odometry, LCD, and the back-end robust estimator. The cross-robot localization topic is closely related to crowd-sourced mapping~\cite{herb2019crowd} for self-driving cars, which involves other important topics that are beyond the scope of this survey.

\section{Open Problems}
\label{sec:discussion}

We begin the discussion section with a question: which is the best global LiDAR localization method? We consider that it is decided by many key factors: environments, maps and required pose accuracy, etc. There is no single best method to handle all applications and scenarios. \textit{Users need to customize the global localization system according to what they actually need.} For example, particle filter family~\cite{dellaert1999monte} is widely used to handle 3-DoF re-localization on wheeled robots. For vehicle localization in urban environments, scan context family~\cite{kim2018scan} could be a good choice for loop closure detection due to its simplicity and learning-free scheme. For local pose estimation in SLAM, we kindly take MULLS~\cite{pan2021mulls} as an example. MULLS extracts feature points and uses TEASER~\cite{TeaserTRO2021} for global feature registration in its loop closure module. Overall, modern global LiDAR localization techniques have enabled several important functionalities for mobile robots. However, there are still open problems and worthy topics for future study. We will discuss these problems and conclude several promising directions for global LiDAR localization.

\begin{table*}[t]
\centering
\caption{\textbf{Public Datasets for Global LiDAR Localization.}}
\label{tab:datasets}
\begin{tabularx}{\textwidth}{cccccccc}
\hline
Dataset & Scenarios & Total Distance & Challenges & \begin{tabular}[c]{@{}c@{}}Viewpoint\\ Diversity\end{tabular} & \begin{tabular}[c]{@{}c@{}}Dynamic \\ Objects\end{tabular} & \begin{tabular}[c]{@{}c@{}}Multi-\\ Session\end{tabular} & \begin{tabular}[c]{@{}c@{}}Sensor\\ (\# fov)\end{tabular} \\ \hline 
KITTI ~\cite{geiger2013vision} & Urban & $\sim$44km & - & $\star$ & $\star$ & $\times$ & full \\
[0.25cm]
NCLT ~\cite{carlevaris2016university} & Campus & $\sim$147km & \begin{tabular}[c]{@{}c@{}}Viewpoint \\ Change\end{tabular} & $\star$$\star$$\star$ & $\star$ & $\checkmark$ & full \\
[0.25cm]
Oxford RobotCar ~\cite{maddern20171} & \begin{tabular}[c]{@{}c@{}}Urban\\ + Suburban\end{tabular} & $\sim$1000km & Occlusions & $\star$$\star$ & $\star$$\star$$\star$ & $\checkmark$ & full \\ 
[0.25cm]
Apollo-SouthBay ~\cite{lu2019l3} & Urban & $\sim$381km & - & $\star$ & $\star$$\star$ & $\checkmark$ & full \\
[0.25cm]
\begin{tabular}[c]{@{}c@{}}Oxford \\ Radar RobotCar ~\cite{barnes2020oxford} \end{tabular}  & \begin{tabular}[c]{@{}c@{}}Urban\\ + Suburban\end{tabular} & $\sim$280km & \begin{tabular}[c]{@{}c@{}}Occlusion \\ Cross Modality \end{tabular} & $\star$$\star$ & $\star$$\star$$\star$ & $\checkmark$ & full \\ 
[0.25cm]
MulRan ~\cite{kim2020mulran} & \begin{tabular}[c]{@{}c@{}}Urban\\ + Campus\end{tabular} & $\sim$123km & \begin{tabular}[c]{@{}c@{}} Less Overlap \\ Cross Modality \end{tabular} & $\star$$\star$ & $\star$$\star$ & $\checkmark$ & $270^{\circ}$ \\
[0.25cm]
Newer College ~\cite{ramezani2020newer} & Campus & $\sim$7km & \begin{tabular}[c]{@{}c@{}}Viewpoint \\ Change\end{tabular} & $\star$$\star$$\star$ & $\star$ & $\times$ & full \\
[0.25cm]
KITTI360 ~\cite{liao2022kitti} & \begin{tabular}[c]{@{}c@{}}Urban\\ + Suburban\end{tabular} & $\sim$74km & - & $\star$ & $\star$ & $\times$ & full \\
[0.25cm]
ALITA ~\cite{yin2022alita} & \begin{tabular}[c]{@{}c@{}}Urban\\ + Campus\end{tabular} & $\sim$60/120km & - & $\star$$\star$ & $\star$$\star$ & $\times$ & full \\
[0.25cm]
Wild-Places~\cite{knights2022wild} & Forest & $\sim$33km & \begin{tabular}[c]{@{}c@{}}Viewpoint \\ Change\end{tabular} &  $\star$$\star$$\star$ & N/A & $\checkmark$ & full \\
[0.25cm]
Boreas~\cite{burnett_ijrr23} & Urban & $\sim$350km & \begin{tabular}[c]{@{}c@{}}Occlusion \\ Cross Modality\end{tabular} &  $\star$ & $\star$$\star$$\star$ & $\checkmark$ & full \\
[0.25cm]
HeLiPR ~\cite{jung2023helipr}  & Urban & $\sim$145km & Less Overlap &  $\star$$\star$ & $\star$$\star$$\star$ & $\checkmark$ &  \begin{tabular}[c]{@{}c@{}}$70^{\circ}$/$120^{\circ}$/ \\ full\end{tabular} \\
[0.1cm]
\hline
\end{tabularx}
\end{table*}


\subsection{Evaluation Difference}
\label{sec:evaluation}

Experimental validation and evaluation are critical for research papers. We notice that related papers evaluate their methods with different metrics and datasets. In the following subsections, we first list the commonly used metrics and then discuss the evaluations.


\textbf{Place Recognition Metrics.} One branch of metrics is for evaluating the place retrieval performance. Any place recognition methods evaluate the global localization performance under machine learning metrics, like Recall@1\%~\cite{uy2018pointnetvlad}, precision-recall curves~\cite{kong2020semantic} and localization probability~\cite{dube2017segmatch,yin20193d} etc. In this context, a robot is localized successfully if the retrieved place is close to the ground truth position (\textless$d$ m). The threshold $d$ is a user-defined parameter and related to the resolution of topological keyframe-based submaps, like 25m in~\cite{uy2018pointnetvlad} and 3m in~\cite{kong2020semantic}. Popular metrics are illustrated as follows.

\begin{enumerate}
    \item True Positives TP, False Positives FP, and False Negatives FN
        \begin{itemize}
            \item \textit{True Positives TP} is the number of actually matched places that are recognized as matched places.
            \item \textit{False Positive FP} is the number of actually not matched places that are recognized as matched places.
            \item \textit{False Negatives FN} is the number of actually matched places that are recognized as not matched places.
        \end{itemize}
        
    \item Precision and Recall
        
        \textit{Precision} and \textit{Recall} are defined based on \textit{TP}, \textit{FP} and \textit{FN}. \textit{Precision} represents the ratio of true positives to total queries, calculated as
        \begin{equation}
        Precision = \frac{TP}{TP + FP}
        \end{equation}
        \textit{Recall} represents the ratio of true positives to total positives, formulated as
        \begin{equation}
        Recall = \frac{TP}{TP + FN}
        \end{equation}

    \item Precision-recall Curve
    
    Instead of using a single threshold for evaluation, \textit{Precision-recall Curve} depicts \textit{Precision} as a function of \textit{Recall} $Precision = f(Recall)$ at different thresholds.
    \item F1 Score and AUC

     \textit{F1 Score} is a combined metric of \textit{Precision} and \textit{Recall}, which is the harmonic mean them, expressed as
     \begin{equation}
     F1 \; Score = \frac{2 \times Precision \times Recall}{Precision + Recall}
     \end{equation}
     \textit{AUC} is the area under \textit{Precision-recall Curve}, compressing \textit{Precision-recall Curve} into a single value.  
     \item Recall@N and Recall@1\%
     
     As the first step of global localization, place recognition algorithms usually retrieve the top-N matched places in the database for each query for further geometric verification, where \textit{Recall@N and Recall@1\%} are more suitable for performance evaluation, formulated as
    \begin{equation}
    Recall@N = \frac{TP_{topN}}{TP_{topN} + FN_{topN}},
    \end{equation}
    \begin{equation}
    Recall@1\% = \frac{TP_{top1\%}}{TP_{top1\%} + FN_{top1\%}}
    \end{equation}     
    where \textit{Recall@N} equals to \textit{Recall@1\%} if $N$ equals one percent of the total number of places in the database.
\end{enumerate}

\textbf{Pose Estimation Metrics.} The other evaluation metrics are designed for pose estimation. Conventional LiDAR SLAM and map-based localization evaluate the performance based on rotation and translation errors. These errors are obtained by comparing the estimated global pose with the ground truth pose quantitatively. Several global localization approaches follow these metrics~\cite{kim2021scan,cattaneo2022lcdnet}.

\begin{enumerate}
    \item Translation Error TE and Rotation Error RE
    
    \textit{TE} is the error between the estimated translation and the ground-truth translation, which is calculated by
    \begin{equation}
    TE = \Vert \hat{d}-d\Vert
    \end{equation}
    \textit{RE} is the error between the estimated rotation and the ground-truth rotation, whose mathematical formula is
    \begin{equation}
    RE = \arccos{((trace(\hat{R}^TR)-1)/2)}
    \end{equation}  
    where $\hat{d}$ and $\hat{R}$ are the estimated translation vector and rotation matrix. $d$ and $R$ are the ground-truth translation vector and rotation matrix.

    \item Successful Rate

    \textit{Successful Rate} shows the ratio of successfully localized cases to total cases. Regarding the localization results with $TE < \tau_{TE}$ and $RE < \tau_{RE}$ as success, \textit{Successful Rate} is defined with 
    \begin{equation}
    Successful \; Rate = \frac{TP_{TE < \tau_{TE} \; \& \; RE < \tau_{RE}}}{TP + FP}
    \end{equation}
    where $\tau_{TE}$ and $\tau_{RE}$ are the thresholds of \textit{TE} and \textit{RE}.
    
\end{enumerate}





\textbf{Metric Difference.} For place recognition, one might ask which metric is the best choice to evaluate the place recognition performance. We consider this depends on the task in practical situations, as we discussed in Section~\ref{sec:situations}. For instance, the LCD provides the edges for the pose graph optimization, and a false positive may ruin the back-end optimization. Consequently, a high level of \textit{Precision} might be desired for certain LCD tasks. In terms of re-localization, the \textit{Recall} can show how well the system can localize a robot from an initial state. 

On the other hand, LiDAR sensors provide precise and stable range measurements compared to other onboard sensors. More importantly, in a classical scheme of autonomous navigation~\cite{siegwart2011introduction}, an accurate pose state is desired from downstream planning and control modules. Hence, we argue that \textit{place retrieval is not the ultimate goal for the global LiDAR localization problem, and pose estimation metrics are more meaningful.} As presented in the previous review~\cite{toft2020long}, visual localization methods are evaluated and discussed with 6-DoF poses. Currently, there is no lidar localization evaluated in the same pose estimation metric as the one in~\cite{toft2020long}, and it could be a future direction for the study on global LiDAR localization. Despite the discussion above, we want to emphasize that the pose estimation should be task-dependent. For a planar moving robot, 3-DoF pose estimation could be enough for LCD and re-localization, or it could be an initial guess for 6-DoF estimation if needed. We recommend readers evaluate global localization based on practical situations in the real world.

\textbf{Long-term Evaluation.} Another important and meaningful topic is the evaluation of long-term global localization. Studies and findings have been reported in both visual and LiDAR place recognition papers~\cite{alijani2022long,peltomaki2021evaluation}. For place-recognition-based approaches, we consider conventional metrics may not be sufficient for performance evaluation as they mainly focus on short-term retrieval performance. Pioneering works~\cite{cui2023ccl,knights2022incloud} have proposed solutions for continual learning in LiDAR-based place recognition across diverse seasons and cities. They introduce a new ``Forgetting'' score, quantifying the extent to which the model forgets past learnings after being trained on data from new environments. However, the "Forgetting" score is derived from precision-recall results provided by place recognition, which may not be suitable for evaluating metric global localization methods. Designing novel evaluation metrics for long-term global localization remains an open challenge.

\textbf{Public Datasets.} Lastly, Table~\ref{tab:datasets} is listed to help researchers grasp representative datasets for global LiDAR localization. We summarized and evaluated these datasets from multiple perspectives, such as scenarios, challenges and viewpoint diversity. The challenges and relevant conditions in the table will be introduced in the following sections. Generally, different datasets cover experimental conditions and challenges, for example, Boreas~\cite{burnett_ijrr23} was collected in multi-session urban environments and it includes both LiDAR and radar as range sensings. Thus Boreas is an ideal dataset to evaluate long-term and cross-modality global localization in urban scenes. We recommend users select suitable datasets for performance evaluation.

\subsection{Multiple Modalities}
Modern mobile robots are equipped with multiple sensors for self-localization~\cite{jiao2022fusionportable}. In recent years, multi-modal sensing has been a hot topic in the community and has attracted much research interest. Different modalities bring direct challenges for \textit{cross-modality} global localization. But on the other hand, each sensor modality has its pros and cons, and sensor \textit{modality fusion} can potentially improve the reliability and robustness of localization. Despite the sensor modalities at the front end, recent learning techniques enable modality study to higher-level tasks, and we will also introduce \textit{high-level semantics} in the LiDAR global localization problem.

\noindent \textbf{Cross Modality.} When offline mapping and online localization use different sensor modalities, we name it cross-modality localization. Cattaneo \emph{et al.}~\cite{cattaneo2020global} train 2D images and 3D point place recognition jointly. To achieve this, a deep neural network is built, integrating classical 2D convolution layers and 3D PointNet. Similarly, in~\cite{yin2021radar}, radar and LiDAR are mixed for BEV-based place recognition. Overhead satellite imagery is a cheap source for outdoor localization that does not require lidar mapping beforehand. Metric global LiDAR localization on 2D satellite imagery is proposed and validated in~\cite{tang2021get}. OpenStreetMap (OSM) is also an alternative map that includes structural road and building information. In~\cite{cho2022openstreetmap}, handcrafted LiDAR descriptors are matched to OSM descriptors to achieve place recognition. For metric localization, a 4-bit representation is proposed in~\cite{yan2019global} that can measure the hamming distance between laser scans and OSM. Then the distances are used to formulate the observation model of MCL. Overall, the research insight of these works~\cite{cattaneo2020global,yin2021radar,tang2021get,cho2022openstreetmap,yan2019global} is to build a shared low-dimensional representation that can connect different data modalities. Learning techniques are needed to extract the shared feature embeddings in most of these existing approaches. We consider precise and global cross-modality global localization is still a challenging problem, e.g., matching a 2D image on a 3D point map globally.

\noindent \textbf{Modality Fusion.} Another direction is to build sensor fusion modules based on multiple modalities. A LiDAR-Vision segment descriptor achieves better performance for place recognition tasks than a LiDAR-only descriptor, as validated in~\cite{ratz2020oneshot}. Inspired by this, Coral~\cite{pan2021coral} designs a bi-modal place recognition by fusing colorful visual features and structural LiDAR elevation maps. AdaFusion~\cite{lai2022adafusion} uses an attention scheme to weight visual and LiDAR modalities for place recognition. In~\cite{bernreiter2021spherical}, a spherical projection enables the fusion of visual and LiDAR at the front end without losing information. From these works above~\cite{ratz2020oneshot,pan2021coral,lai2022adafusion,bernreiter2021spherical}, we can conclude that modality fusion can help improve the place recognition performance, but extra learning techniques or training data are needed to fuse different modalities.

\noindent \textbf{High-level Semantics} As mentioned in Section~\ref{sec:PR}, raw LiDAR point clouds are textureless and in an irregular format compared to visual images. This constrains high-level robotics applications, such as scene understanding and moving object detection. Behley \emph{et al.}~\cite{behley2021ijrr} released a large semantic LiDAR dataset in 2019, named SemanticKITTI. SemanticKITTI contains point-wise annotated LiDAR scans, and multiple semantic-related benchmarks for on-road autonomous navigation. A backbone network RangeNet++~\cite{milioto2019rangenet++} and a semantic LiDAR SLAM SuMa++~\cite{chen2019suma} are also released publicly trained on semantic information provided by SemanticKITTI. The SemanticKITTI focuses on on-road robotic perception tasks, and we also note that there exists an off-road semantic LiDAR dataset, named RELLIS-3D~\cite{jiang2021rellis}. RELLIS-3D also provides a full stack of multi-modal sensor data for field robotics research. The semantic information of LiDAR benefits both place recognition and pose estimation for global localization. In~\cite{kong2020semantic,vidanapathirana2021locus,pramatarov2022boxgraph}, semantics are used to construct a semantic-spatial graph for global descriptor extraction. It is validated that using semantic information can improve place recognition performance under this graph representation. Semantic information is also used to improve the performance of LiDAR odometry~\cite{chen2019suma} and global point cloud registration~\cite{li2021ssc,yin2023segregator}. Beyond the point-level semantics, recent study~\cite{xia2023text2loc} also integrates natural language descriptors into the LiDAR place recognition, making the robot semantically understand the environments.

\subsection{Less Overlap}
\label{sec:lessoverlap}
Though the LiDAR scanner is powerful for environmental sensing, there exists a potential challenge when applying global LiDAR localization in practice: the overlap between two LiDAR point clouds might be very small in certain cases. For the global localization problem, the point clouds could be two scans or submaps for place retrieval, or point cloud registration for pose estimation. Less overlap will make place recognition or pose estimation techniques much more challenging, where there emerge some works try to tackle it in the recent two years~\cite{liu2023regformer,qiao2023g3reg}. To better understand this challenge, we list three typical cases as follows.

\noindent \textbf{Occlusions by Dynamics.} LiDAR scans will be partially blocked by dynamics in a high-dynamical environment, like pedestrians and vehicles around the robot. More specifically, for a spinning LiDAR sensor, the block area is decided by mainly two factors: the distance between dynamics and sensor, and the size of this dynamic. By far, dynamics removal in LiDAR scans effectively and efficiently is still quite difficult~\cite{lim2021erasor}. Compared to 360-degree rotating lidar sensors, some other range sensors can provide range data that are not easily blocked, like imaging radar~\cite{kramer2022coloradar}, and spinning radar~\cite{kim2020mulran}.

\noindent \textbf{Large Translation.} For sparse keyframe-based submaps, a large translation between the retrieved keyframe and ground truth pose could result in a small overlap between the current LiDAR scan and the submaps stored in the keyframe. A powerful global point cloud registration is required to overcome this challenge.

\noindent \textbf{Viewpoint Change.} Generally, for wheeled robots on roads, pose estimation is constrained in a 3-DoF space (x, y and yaw). But for flying drones in the wild, it is a complete 6-DoF pose estimation problem. When using global LiDAR localization on drones, LiDAR point clouds collected by drones might have less overlap at the same place. This is mainly due to two reasons: the 6-DoF motions of drones and the limited field-of-view (FoV) of LiDAR sensors.

\subsection{Unbalanced Matching}
\label{sec:unbalance}

Most global LiDAR localization methods are validated using relatively good data quality. In other words, the input point cloud and point cloud map have similar distributions and the same representations. However, in practical situations, the input and the map are usually unbalanced. We list three typical considerations as follows.

\textbf{Scan to Submap.} Keyframe-based map is a popular map organization in large-scale scenes, as introduced in Section~\ref{sec:KM}. Matching a single scan to a submap globally is a crucial step to re-localize a mobile robot when localization fails. However, LiDAR point cloud submaps are generally with larger scales compared to single LiDAR scans, and also with a density variance. Traditional matching methods need tuned parameters and thresholds to handle these cases. For learning-based approaches, unbalanced point matching~\cite{lee2022learning} is still hard to solve since local features are quite different~\cite{chang2021map} in such conditions.

\textbf{Representation Difference.} There exist other metric map representations beyond point clouds, aforementioned in Section~\ref{sec:GMM}. Generally, multiple representations are integrated into the navigation paradigm, e.g., localization in point cloud maps and planning on grid maps. To simplify the map use, one potential direction is to build a unified representation for navigation and unbalanced matching is needed to overcome the representation difference.

\textbf{Noisy LiDAR.} LiDAR sensors will be affected in challenging conditions, like rainy and snowy days~\cite{pitropov2021canadian} and even strong lights on roads~\cite{carballo2020libre}, resulting in noises in raw LiDAR data. These noises directly bring hazards for robot perception and state estimation. Dealing with the noises is important to guarantee the safety and robustness of the navigation system.

\subsection{Efficiency and Scalability}
\label{sec:efficiency}

Efficiency and scalability are significant considerations in LiDAR-based localization tasks, as quick and accurate processing of incoming LiDAR data is essential for timely decision-making in applications like autonomous vehicles. However, current approaches have not efficiently tackled LiDAR-based localization in large-scale environments, especially on incrementally enlarged city-scale maps. One reason is the inherent characteristics of LiDAR data, which is often large and high-dimensional. Processing it in real-time can be computationally demanding, and handling large-scale environments with high-density point clouds requires efficient algorithms and hardware. On the other hand, the generation and real-time updating of accurate maps from LiDAR data for localization purposes are complex, particularly in dynamically changing large-scale environments. A widely adopted tool for organizing databases in key-frame-based methods and facilitating the location retrieval process has been introduced by Facebook Research named Faiss~\cite{johnson2019billion}. Faiss is built around an index type that stores a set of keyframe descriptors and provides a function to search in them efficiently using the fast k-means with GPU. Although Faiss is effective for descriptor-based approaches, achieving efficient global localization for other map-type-based methods remains an open challenge.

\indent Compressing the point cloud~\cite{wiesmann2021deep} could be a promising way to reduce the demand on LiDAR map storage of large-scale environments. However, the current approach~\cite{wiesmann2022retriever} needs an extra decompression step when using such compressed maps for 6-DoF localization, making it a trade-off between storage and speed.

\subsection{Generalization Ability}
\label{sec:generalization}

For learning-free methods, less parameter tuning is desired to ensure the generalization to new environments~\cite{vizzo2022kiss}. As for learning-based methods, generalization ability is a big challenge that has to face, especially when there is less training to support these data-driven methods. We mainly list four basic problems when deploying existing global localization methods.

\noindent \textbf{Sensor Configuration.} Currently, there are dozens of LiDAR types, and each type has its unique sensor parameters. The generalization from one LiDAR sensor to another could be a problem, e.g., training on Velodyne HDL-64E scans while testing on Ouster OS1-128 scans. Another potential problem is the displacement of LiDAR sensors. If roll or pitch angle changes, laser point density and distribution will change respectively, resulting in global localization failure even using state-of-the-art methods. However, if the global localization is conducted on accumulated submaps, sensor configuration could be a minor problem.


\noindent \textbf{Unseen Environments.} The generalization in the unseen scenario is an old but still hard problem in the learning community. Cross-city and Cross-environments generalization remains underexposed for global LiDAR localization methods. For instance, Knights \emph{et al.}~\cite{knights2022wild} release a challenging dataset Wild-Places for LiDAR place recognition in natural environments. There is a performance drop for advanced methods~\cite{kim2018scan,xu2021transloc3d,komorowski2022improving} compared to tests in urban environments. It could be concluded that there is a domain gap between structural urban environments and unstructured natural environments. To enable continuous learning in new scenes, incremental learning~\cite{li2017learning} is a good choice that does not require retraining from scratch. Recent work InCloud~\cite{knights2022incloud} achieves incremental learning for point cloud place recognition and it overcomes catastrophic forgetting caused by learning in new domains.

\noindent \textbf{Trigger of Global Localization.} In a complete localization system, pose tracking takes most of the computation while global localization is only activated when it is needed. Thus, a natural question is raised: when to trigger global localization? For LCD and cross-robot localization, the trigger of global localization could be one or multiple pre-defined criteria, like similarity threshold of descriptors or an adaptive distance in~\cite{denniston2022loop}. As for re-localization applications, a robot might believe it knows where it is while it does, and it is actually the classic kidnapped robot problem in~\cite{thrun2002probabilistic}. In this context, detected localization failure could be a trigger condition for global re-localization. The LiDAR localization failure detection problem is identical to the point cloud registration quality evaluation at the front end. Researchers propose to design multiple metrics and train classifiers to learn how to score this registration quality~\cite{yin2019failure,adolfsson2022coral}. While at the back-end, features of state estimator can be used for failure detection~\cite{fujii2015detection}. Localization failure prediction and avoidance is also a worthwhile studied topic for long-term autonomy~\cite{nobili2018predicting}.

\section{Conclusion}
\label{sec:conclusion}



The field of global localization has attracted significant attention from researchers in recent years, due to its pivotal role in mobile robot applications. The increasing number of innovative studies with LiDAR sensors has motivated us to organize a comprehensive survey on the global LiDAR localization problem. This survey aims to aggregate existing advanced knowledge, while simultaneously identifying problems for future research.

Our review starts with the problem formulation from a probabilistic view. Then we consider typical situations in real-world applications: loop closure detection, re-localization, and cross-robot localization. This initial analysis provides a foundation for understanding the problem and positions the various methodologies within their appropriate application scopes. Then the structure of the contents is organized into three themes. The first theme delves into global place retrieval and local pose estimation, exploring how these two concepts interact within the broader context of global localization. The subsequent theme presents an evolution from single-shot measurements to sequential ones, emphasizing how this progression enhances sequential global localization. The final theme broadens the scope to consider the extension of single-robot global localization to cross-robot localization in multi-robot systems, highlighting the complexities and opportunities in this emerging area.

One might ask whether this problem is solved or not. We consider there are still many promising research directions, as we discussed in Section~\ref{sec:discussion}. In addition to the problem itself, the integration of global localization into the navigation system also represents a valuable research topic. This involves examining the system architecture and its operating environments. Given that robotics is often a case-specific study, we recommend users tailor the global localization to their requirements.

\section*{Acknowledgment}

 We would like to thank Dr. Xiaqing Ding for her constructive suggestions. 

\bibliographystyle{unsrt}
\footnotesize
\bibliography{main}

\end{sloppypar}

\end{document}